\documentclass[11pt]{article}

\usepackage[preprint]{acl}
\usepackage{conference_ACL}
 \usepackage[table]{xcolor}
 \usepackage{booktabs}
 \usepackage{multirow}
 \usepackage{subfig}

 \newcommand{\qcol}{\cellcolor{blue!5}}
\title{LPCD: Unified Framework from Layer-Wise to Submodule Quantization}

\author{Yuma Ichikawa \\
  Fujitsu Limited \\
  RIKEN Center for AIP \\ 
  \And
  Yudai Fujimoto \\
  Fujitsu Limited \\
  Institute of Science Tokyo \\ 
  \And
  Akira Sakai \\
  Fujitsu Limited \\
  Tokai University  \\
  \\}

\begin{document}
\maketitle 
\begin{abstract}
    Post-training quantization (PTQ) aims to preserve model-level behavior; however, most methods focus on individual linear layers.
    Even recent extensions, such as QEP and LoaQ, which mitigate error propagation or target specific submodules, still rely on layer-wise formulations and fail to capture the behavior of larger submodules.
    We introduce \underline{\textbf{L}}ayer-\underline{\textbf{P}}rojected \underline{\textbf{C}}oordinate \underline{\textbf{D}}escent (\textbf{LPCD}), a unified framework that extends PTQ beyond layers by optimizing relaxed objectives across arbitrary submodules and projecting the solutions with standard layer-wise quantizers.
    LPCD generalizes existing methods and provides a principled approach to quantizing complex submodules while maintaining the efficiency and compatibility of layer-wise PTQ pipelines.
    Across diverse LLM architectures and bit-widths, LPCD-based submodule quantization consistently enhances both layer-wise PTQ methods and existing submodule approaches.
\end{abstract}

\section{Introduction}\label{sec:intro}

\begin{figure*}
  \centering
  \subfloat[4-bit weight]{%
    \includegraphics[width=0.32\linewidth]{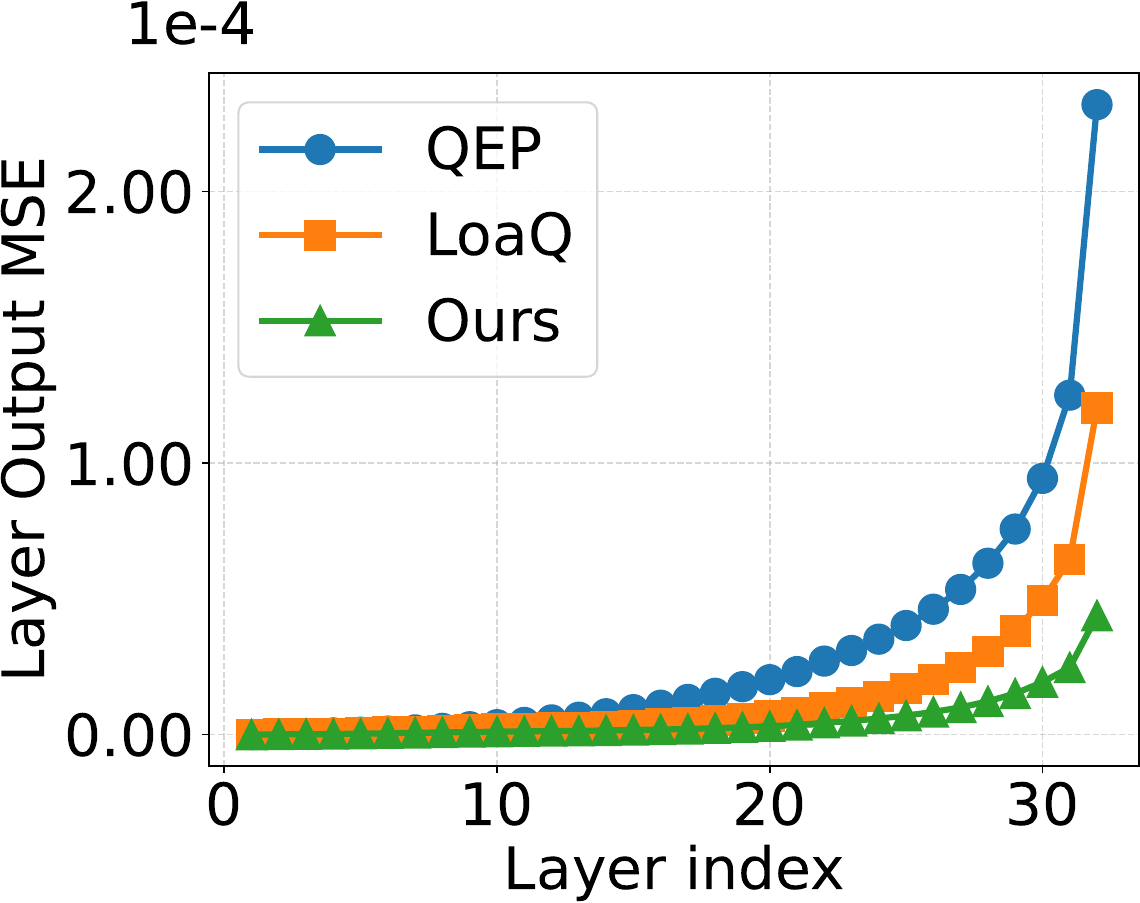}
    \label{fig:sub1}
  }
  \hfill
  \subfloat[3-bit weight]{%
    \includegraphics[width=0.32\linewidth]{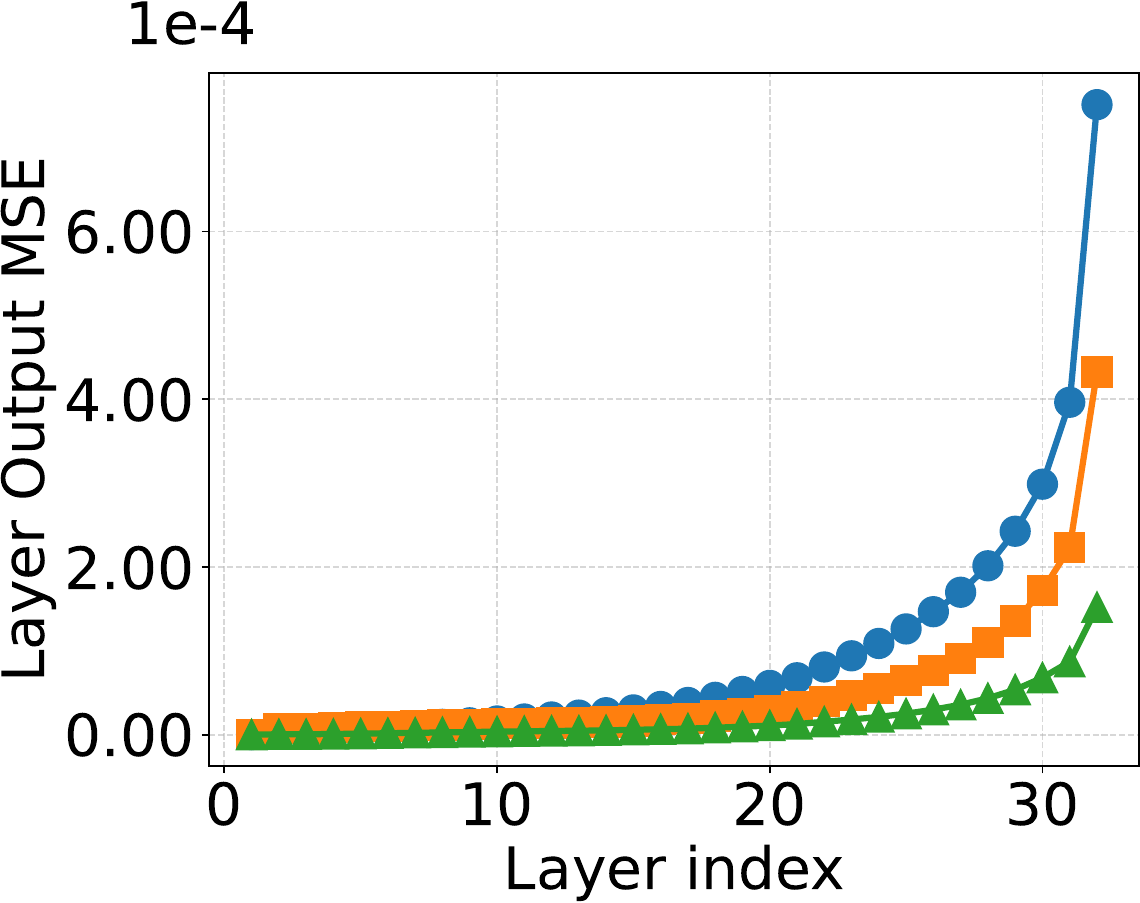}
    \label{fig:sub2}
  }
  \hfill
  \subfloat[2-bit weight]{%
    \includegraphics[width=0.32\linewidth]{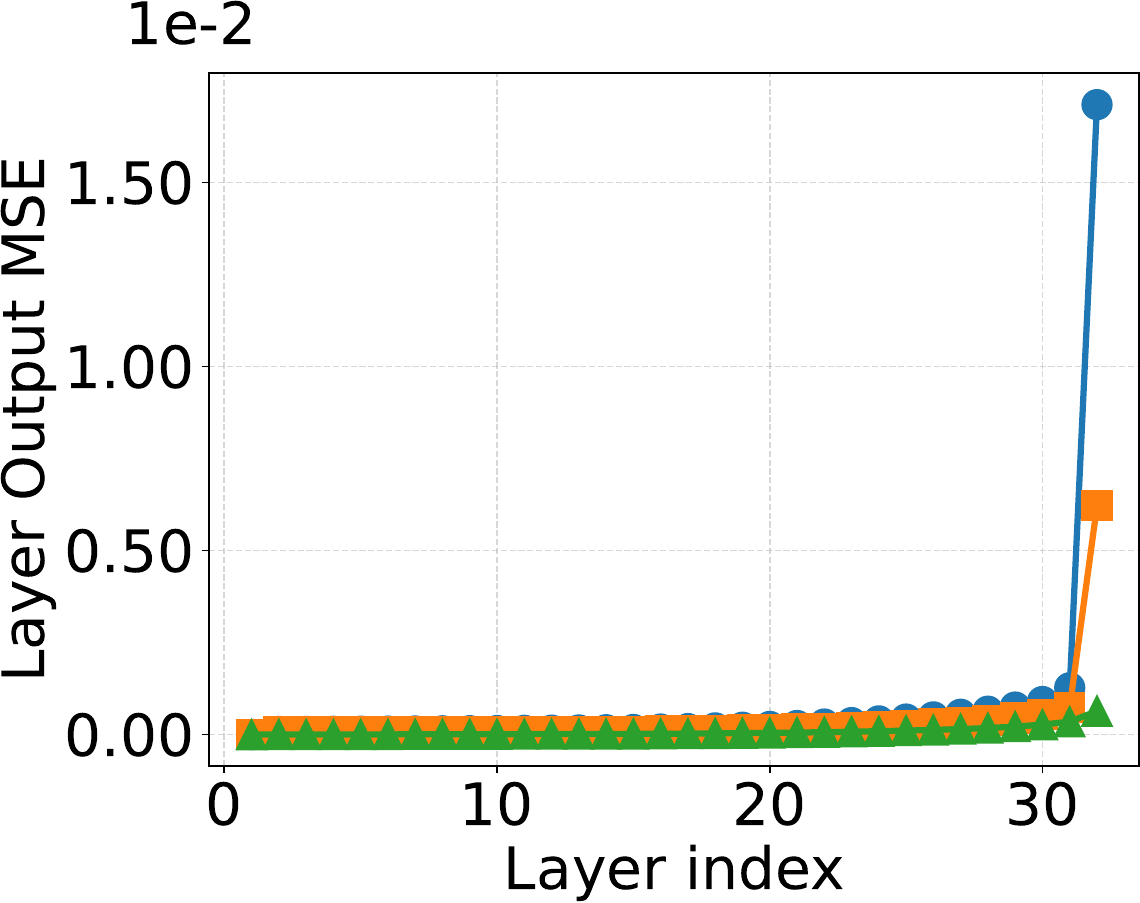}
    \label{fig:sub3}
  }
  \caption{Output MSE across Transformer blocks in Llama~3~8B with 4, 3, and 2 bit weight quantization. 
  LPCD consistently yields lower quantization error than QEP and LoaQ.}
  \label{fig:whole-main-fig}
\end{figure*}

Large-scale models achieve strong performance; however, they incur substantial memory and computational overhead, hindering practical deployment~\citep{chen2023frugalgpt}.
These constraints are particularly stringent for edge devices.
To bridge the gap between accuracy and deployability, prior work has explored compression techniques such as quantization~\citep{lang2024comprehensive, gong2024survey}, pruning~\citep{wang2024model, cheng2024survey}, low-rank adaptation~\citep{yang2024low, hu2022lora}, and knowledge distillation~\citep{xu2024survey}.
Among these techniques, layer-wise post-training quantization (PTQ) is one of the most practical and widely adopted methods for large-scale LLMs~\citep{frantar2022gptq, lin2024awq, yao2022zeroquant, chee2023quip}.
By focusing on individual linear layers, layer-wise PTQ simplifies the problem to least-squares estimation of linear transformations, which enables for efficient solvers and straightforward implementations.
Despite its simplicity, layer-wise PTQ delivers strong empirical performance and is used both as a standalone method and as an initialization for more complex block-wise or global PTQ frameworks~\citep{malinovskii2024pv, guan2024aptq}.

However, the layer-wise structure imposes significant limitations.
Classical layer-wise PTQ methods, such as GPTQ~\citep{frantar2022gptq}, AWQ~\citep{lin2024awq}, and QuIP~\citep{chee2023quip}, optimize each linear layer independently by utilizing input activations.
Consequently, these methods are confined to activation-aware weight approximation and indirectly affect model-level outputs; quantization error may become significant.
Recent work, including QEP~\citep{arai2025quantization} and GPTAQ~\citep{li2025gptaq}, relaxes this constraint by modifying the layer-wise objective to address the mismatch between pre- and post-quantization activations. 
This modification enables controlled error propagation across layers, preserving the sequential pipeline and reducing the accumulation of quantization errors.
LoaQ~\citep{lin2025loaq} further extends the target from linear layers to specific submodules by explicitly approximating the outputs of residual connections and RMSNorm. 
This approach aligns layer-wise PTQ more closely with the behavior at the model-level.
Nevertheless, these advances remain specialized: QEP is still anchored in individual linear layers, and LoaQ is limited to specific submodules; they do not provide a unified treatment of more general submodules, activation, or KV-cache quantization.

This study goes beyond traditional layer-wise formulations by introducing a framework for submodule quantization that preserves the standard layer-wise pipeline.
We propose \textbf{Layer-Projected Coordinate Descent (LPCD)}, which optimizes arbitrary submodules in the output space and projects the relaxed solutions back into the quantization domain using existing layer-wise PTQ algorithms as \emph{layer projectors}.
This perspective unifies classical layer-wise PTQ, QEP, and LoaQ as special cases while extending naturally to more general submodules, activations, and KV caches.
We instantiate LPCD on grouped-query KV, VO aggregation, and MLP up-down blocks, demonstrating that LPCD significantly reduces quantization error compared to QEP and LoaQ, as shown in Figure~\ref{fig:whole-main-fig}.
Extensive experiments across diverse tasks and bit-widths demonstrate that LPCD-based submodule quantization consistently enhances both layer-wise PTQ methods and existing submodule approaches. Our main contributions are as follows:
\begin{itemize}
    \item LPCD is a \emph{unified framework} that extends layer-wise PTQ to \emph{arbitrary} submodules by reusing existing layer-wise quantizers.
    \item An effective method for quantizing coherent Transformer submodules reduces quantization error while remaining fully compatible with layer-wise PTQ pipelines.
    \item LPCD consistently enhances both layer-wise PTQ and existing submodule methods across diverse bit-widths and models, while avoiding unstable STE heuristics \citep{bengio2013estimating}.
\end{itemize}

\section{Related Work}\label{sec:related-work}

Quantization for LLMs includes data-free PTQ, layer-wise PTQ, block or global PTQ, and QAT~\citep{dettmers2023case, frantar2022gptq, tseng2024quip, egiazarian2024extreme, xu2024onebit, wang2023bitnet, liu2023llm}.
Among these, weight-only layer-wise PTQ is particularly popular because each linear layer reduces to a least-squares problem that can be efficiently quantized on a single GPU~\citep{frantar2022gptq, lin2024awq, chee2023quip, zhao2025benchmarking}.
Layer-wise PTQ methods are typically implemented as \emph{compensation-based} schemes, such as GPTQ and its variants~\citep{frantar2022gptq, behdin2023quantease, liu2024vptq, guan2024aptq}, \emph{rotation-based} schemes, such as QuIP and its successors~\citep{chee2023quip, tseng2024quip, liu2024spinquant, ashkboos2024quarot}, or \emph{salience-based} schemes, such as AWQ and mixed-precision methods~\citep{dettmers2022gpt3, dettmers2023spqr, shang2023pb, lin2024awq}.
More recently, QEP \citep{arai2025quantization}, GPTAQ \citep{li2025gptaq}, and Qronos \citep{zhang2025qronos} refine the \emph{layer-wise loss} by compensating for mismatches between pre- and post-quantization activations at the linear-layer level. Meanwhile, LoaQ extends this concept to sub-layers, including residual connections and RMSNorm~\citep{lin2025loaq}.
Our work follows this direction but provides a unified framework that applies layer-wise PTQ operators to minimize losses over more general submodules, such as KV, VO, and MLP, extending beyond individual layers or a fixed set of sub-layers.

\section{Preliminaries}\label{sec:preliminaries}

\subsection{Notation}
In this section, we introduce the notation used consistently throughout the paper.
Let $[R] \coloneqq \{1, \ldots, R\}$ be a fixed finite index set. 
The expression $\|\cdot \|_{F}$ denotes the Frobenius norm.
The set $\mathbb{Q} \subset \mathbb{R}$ denotes a fixed finite set of quantization levels, e.g., the grid induced by a $b$-bit quantizer, which we refer to as a $b$-bit quantization scheme.
Let $W \in \mab{R}^{N \times M}$ and $\widehat{W} \in \mab{Q}^{N \times M}$ denote the original and quantized weight matrices.
similarly, let $X \in \mab{R}^{T \times N}$ and $\widehat{X} \in \mab{R}^{T \times N}$ represent the activation matrices before and after quantization, where $T$ is the number of tokens and $N$ is the feature dimension.
$R \in \mab{R}^{T \times M}$ and $\widehat{R} \in \mab{R}^{T \times M}$ denote the full-precision and quantized
residual streams, respectively.
We define the following $N \times N$ matrices that arise in the layer-wise quantization loss: $H=X^{\top} X$ and $\widehat{H}=\widehat{X}^{\top} \widehat{X}$, which represent the Hessian matrix of quantization error; $I_{N}$ the $N \times N$ identity matrix.

\subsection{Layer-wise PTQ}

In PTQ, it is commonly assumed that accurately approximating the weights of each linear layer is sufficient to preserve overall model performance; thus, layer-wise PTQ treats each layer as an independent optimization problem.
Layer-wise PTQ typically quantizes layers sequentially in a single forward pass, storing only the inputs of the current layer, which significantly reduces memory usage and computational demand.
Unlike block- or model-wise schemes that must manage nonlinearities and rely on pseudo-gradients, e.g., STE \citep{bengio2013estimating}, layer-wise PTQ circumvents these issues and is consequently often employed as an initializer for more comprehensive methods.
Consequently, the final performance of most PTQ pipelines is largely determined by the quality of the underlying layer-wise quantization.
In the weight-only setting, classical layer-wise objectives decompose into two independent subproblems.

\paragraph{Direct Weight Quantization.}
Direct weight approximation quantizes the model parameters $W$ through the following projection:
\begin{equation}
    \label{eq:direct-original-layerwise-general}
    \widehat{W} = \Pi_{\mab{Q}}^{(d)}(W) \coloneqq \argmin_{\widehat{W}}\|\widehat{W}-W\|_{F}^{2},
\end{equation}
where the objective seeks the optimal quantized value for each weight independently.
A representative method within this framework is the Round-To-Nearest (RTN) technique, which quantizes each weight by rounding it to the closest point in the quantization grid.

\paragraph{Activation-Aware Quantization.}
Activation-aware quantization quantizes the model parameters $W$ by addressing the following activation-aware, layer-wise optimization problem:
\begin{equation}
    \label{eq:activation-aware-layerwise-general}
    \widehat{W} = \Pi_{\mab{Q}}^{(a)}(W) \coloneqq \argmin_{\widehat{W}} \| \widehat{X}(\widehat{W}-W) \|_{F}^{2},
\end{equation}
where the objective explicitly incorporates the input activation distribution, thereby reducing the effect of quantization error on the output of the linear layer.
A typical approach in this category is GPTQ, which leverages the second-order information $H$ of this objective, precomputed and cached for reuse, to further reduce output distortion during quantization.
Leading layer-wise PTQ methods employ distinct optimization strategies to minimize Eq.~\eqref{eq:activation-aware-layerwise-general}.
For example, GPTQ quantizes parameters row-wise by sequentially minimizing reconstruction error and propagating residual corrections to the remaining unquantized entries until each row is fully quantized.
AWQ~\citep{lin2024awq} identifies a small subset of \emph{salient weights} whose magnitudes significantly influence the layer outputs and rescales these weights before quantization.

\subsection{Quantization Error Propagation}
Recent work demonstrates that the basic layer-wise PTQ formulation can be improved by explicitly considering the discrepancy between pre- and post-quantization activations.
QEP is a representative method in this field of study.
Rather than minimizing only the activation-aware loss in Eq.~\eqref{eq:activation-aware-layerwise-general}, QEP directly approximates the outputs of the linear layer by solving
\begin{equation}
    \label{eq:qep-objective}
    \widehat{W}^{\mathrm{QEP}} = \argmin_{\widehat{W}}
    \| \widehat{X}\widehat{W} - XW \|_{F}^{2}.
\end{equation}
As shown in QEP, this objective is equivalent to
\begin{equation}
    \label{eq:qep-transformed-compact}
    \widehat{W}^{\mathrm{QEP}}  = \argmin_{\widehat{W}}
    \|\widehat{X}(\widehat{W} - W^{\ast})\|_{F}^{2}.
\end{equation}
where $W^{\ast}=(I_{N}+\widehat{H}^{-1}C)W$ is accompanied by $C=\widehat{X}^{\top}(X-\widehat{X})$, which represents the error propagation matrix. 
Thus, QEP is first implemented by forming the corrected target $W^{\ast} = (I_{N}+\widehat{H}^{-1}C)W$ and then applying a standard activation-aware layer-wise quantizer to $W^{\ast}$.
To control for overfitting on the calibration set, QEP introduces a tunable coefficient $\alpha \in [0,1]$ and employs $W^{\ast}(\alpha) = (I_{N}+\alpha \widehat{H}^{-1}C)W$, which interpolates between the original
weights $\alpha=0$ and the fully corrected weights.
This simple modification consistently outperforms conventional layer-wise PTQ.

\subsection{Residual Path Quantization}
Beyond individual linear layers, modern Transformers are governed by complex submodules that incorporate multiple linear projections, nonlinear functions, and residual connections.
LoaQ extends the QEP formulation to such submodules, targeting Transformer submodules that consist of a self-attention block, an MLP block, a residual connection, and RMSNorm.
LoaQ directly matches the residual stream by minimizing
\begin{equation}
    \widehat{W}^{\mathrm{LoaQ}}=\argmin_{\widehat{W}}
    \|(\hat{R} + \widehat{X}\widehat{W}) - (R + XW)\|_{F}^{2},
\end{equation}
where $R$ and $\widehat{R}$ denote the full-precision and quantized residual streams, respectively.

As shown in LoaQ, this objective can be expressed in the same activation-aware form as Eq.~\eqref{eq:activation-aware-layerwise-general} with a corrected target $W^\ast(\alpha,\beta)$; the original LoaQ formulation corresponds to using the full corrections with $\alpha=\beta=1$.
\begin{align}
    \label{eq:residual-objective}
    &\widehat{W}^{\mathrm{LoaQ}}= \argmin_{\widehat{W}} \|\widehat{X} (\widehat{W}- W^{\ast}(\alpha, \beta))\|_{F}^{2}, \\
    &W^{\ast}(\alpha, \beta)=(I+\alpha \widehat{H}^{-1}C) W + \beta \widehat{H}^{-1} \Gamma
\end{align}
where $\Gamma = \widehat{X}^{\top}(R - \widehat{R})$ is the residual-path correction and 
$\alpha, \beta \in [0, 1]$ are tunable parameters that enhance stability and prevent over-correction.
Thus, LoaQ replaces the PTQ target with a corrected term that includes both the linear-layer correction $(I_{N}+\widehat{H}^{-1}C)W$ and the residual-path correction $\widehat{H}^{-1}\Gamma$.  
Since standard Transformer blocks apply normalization, such as RMSNorm, after each submodule, the representations used by downstream layers are the \emph{normalized} outputs. 
Accordingly, LoaQ introduces an additional objective that aligns explicitly with these normalized submodule outputs:
\begin{equation}
    \min_{\widehat{W} \in \mab{Q}^{N \times M}}\|\operatorname{Norm}(\hat{R}+\widehat{X}\widehat{W})
          -\operatorname{Norm}(R+XW)\|_{F}^{2},
\end{equation}
where $\operatorname{Norm}(\cdot)$ denotes normalization.
To keep the problem tractable, LoaQ freezes the RMSNorm scaling factors, treating the normalization matrices as fixed, precomputed functions of the inputs so that the dependence on $\widehat{W}$ remains linear. 
Under this mild approximation, LoaQ demonstrates that the aforementioned objective can be rewritten in the same activation-aware form as in Eq.~\eqref{eq:residual-objective}, thereby enabling the direct use of existing layer-wise PTQ solvers while aligning normalized submodule outputs across more complex submodules.

\section{Method}\label{sec:method} 
This section introduces LPCD, a unified framework for quantizing arbitrary submodules that extends layer-wise PTQ beyond these limited scenarios. We first provide a general, architecture-agnostic formulation. We then demonstrate that suitable choices of blocks and losses recover QEP and LoaQ as special cases. Finally, we instantiate LPCD on concrete Transformer submodules, deriving update rules for the QK, VO, and MLP up-down blocks.

\subsection{Problem Setting}\label{subsec:problem-setting}
We consider the following collection of block variables:
\begin{equation}
    \widehat{M}_{r} \in \mab{Q}^{N_{r} \times K_{r}},~~r \in [R].
\end{equation}
These variables are subject to quantization and  represent weight matrices, activation matrices, or, more generally, intermediate model representations such as KV caches. Let
\begin{multline}
    L: \prod_{r=1}^{R} \mab{R}^{N_{r} \times K_{r}} \to \mab{R}, \\
    (M_{1}, \ldots, M_{R}) \mapsto L(M_{1}, \ldots, M_{R})    
\end{multline}
be a loss function that quantifies the discrepancy between a full-precision reference model and its quantized counterpart.
The discrete quantization problem is defined as follows:
\begin{equation}
    \label{eq:global-quantized-M}
    \min_{\widehat{M}_{1},\dots, \widehat{M}_R} L(\widehat{M}_{1},\dots,\widehat{M}_{R}).
\end{equation}
Since $\mab{Q}$ is finite, the feasible set of Eq.~\eqref{eq:global-quantized-M} is a subset of $\prod_{r=1}^{R} \mab{R}^{N_{r} \times K_{r}}$. 
In particular, this feasible set is discrete and non-convex, which makes the direct global optimization of Eq.~\eqref{eq:global-quantized-M} generally intractable.
In this work, we aim to generalize our formulation to encompass both weights and activations by partitioning the index set $[R]$ into two disjoint subsets, $\mac{R}_{w}$ and $\mac{R}_{a}$, such that $\mac{R}_{w} \cup \mac{R}_{a} = [R]$ and $\mac{R}_{w} \cap \mac{R}_{a} = \emptyset$. 
When $r \in \mac{R}_{w}$, the block $M_{r}$ represents the weight matrix of a linear layer, whereas when $r \in \mac{R}_{a}$, the block $M_{r}$ denotes the activation matrix. 

\subsection{Layer-Projected Coordinate Descent}\label{subsec:method-lpcd}

We introduce the LPCD to approximately solve Eq.~\eqref{eq:global-quantized-M}.
For each $r \in [R]$ and outer iteration $t \in\mab{N}$, let $\widehat{M}_{r}^{(t)} \in \mab{Q}^{N_{r} \times K_{r}}$ represent the quantized block $r$ during the $t$-th outer iteration.
We consider a cyclic block-coordinate scheme that updates the blocks in the order $r \in [R]$ and then repeats.
At outer iteration $t$, when updating block $r$, we keep all other blocks fixed at their most recently updated quantized values.
Formally, we define the block-wise objective
\begin{multline}
    L_{r}^{(t)}(U) \coloneqq \\
    L(\widehat{M}_{1}^{(t)}, \ldots, \widehat{M}_{r-1}^{(t)}, U, \widehat{M}_{r+1}^{(t-1)}, \ldots, \widehat{M}_{R}^{(t-1)}),
\end{multline}
where $U \in \mab{R}^{N_{r} \times K_{r}}$ is the relaxed variable corresponding to $\widehat{M}_{r}$ at iteration $t$.
For $s<r$, we use the current-iteration values $\widehat{M}_{s}^{(t)}$; for $s>r$, we use the values from the previous-iteration $\widehat{M}_{s}^{(t-1)}$.
Adapting the sequence of block updates presents an intriguing direction for future research.
The update of block $r \in [R]$ is conducted in two stages.

\paragraph{Relaxation Step.}
We first relax the quantization constraint and solve the corresponding unconstrained continuous optimization problem
\begin{equation}
    \label{eq:relaxation-step}
    \overline{M}_{r}^{(t)} = \argmin_{U} L_{r}^{(t)}(U).
\end{equation}
Whenever $L_{r}$ has a structure that allows for a closed-form minimizer, such as when $L_{r}$ is a strictly convex quadratic function of $U$, we compute the unique minimizer analytically and adopt this as $\overline{M}_{r}^{(t)}$.
Otherwise, we approximate a minimizer, or at least a stationary point of Eq.~\eqref{eq:relaxation-step}, by applying a numerical optimization method, such as gradient descent or an accelerated first-order scheme, to the differentiable function. 
Under standard regularity assumptions, such as the Lipschitz continuity of $\nabla L_{r}$ and suitable step-size conditions, these methods converge to a stationary point of $L_{r}$.
In either case, this step produces a continuous candidate $\overline{M}_{r}^{(t)} \in \mab{R}^{N_{r}\times K_{r}}$ that is either exactly or approximately optimal for block $r$, conditional on the other blocks being fixed at their current quantized values.

\paragraph{Projection Step.}
In this step, we reapply the quantization constraint by projecting $\overline{M}_{r}^{(t)}$ onto $\mab{Q}^{N_{r}\times K_{r}}$ through layer-wise PTQ projections. 
If $r \in \mathcal{R}_{w}$ is a weight matrix, we apply a layer-wise PTQ projection defined in Eq.~\eqref{eq:direct-original-layerwise-general} or Eq.~\eqref{eq:activation-aware-layerwise-general}.
Concretely, we select either the direct weight projection $\Pi_{\mab{Q}}^{(d)}$ or the activation-aware projection $\Pi_{\mab{Q}}^{(a)}$ and define
\begin{equation}
    \label{eq:weight-projection-step}
    \widehat{M}_{r}^{(t+1)}
    =
    \Pi_{\mab{Q}}^{(w)}(\overline{M}_{r}^{(t)}),~\Pi_{\mab{Q}}^{(w)}\in \{\Pi_{\mab{Q}}^{(d)},\Pi_{\mab{Q}}^{(a)}\}.
\end{equation}
If $r \in \mac{R}_{a}$ is an activation matrix, it is quantized by applying the direct projection $\Pi_{\mab{Q}}^{(d)}$ to $\overline{M}_{r}^{(t)}$.
\begin{equation}
    \label{eq:activation-projection-step}
    \widehat{M}_{r}^{(t+1)} = \Pi_{\mab{Q}}^{(d)}(\overline{M}_{r}^{(t)}),~~ r\in\mac{R}_{\mathrm{a}}.
\end{equation}
This choice aims to minimize the entrywise distortion of the Frobenius norm between the full-precision and quantized activations at block $r$.  

By alternating these relaxation and projection steps, each block update maps a feasible quantized tuple $(\widehat{M}_{1}^{(t)}, \ldots, \widehat{M}_{R}^{(t)})$ to another feasible tuple $(\widehat{M}_{1}^{(t+1)}, \ldots, \widehat{M}_{R}^{(t+1)})$ that continues to satisfy the constraints of Eq.~\eqref{eq:global-quantized-M}. 
Since $\mab{Q}$ is finite and the projections are well defined, feasibility is preserved at each iteration. 

\begin{figure*}[tb]
    \centering
    \includegraphics[width=\linewidth]{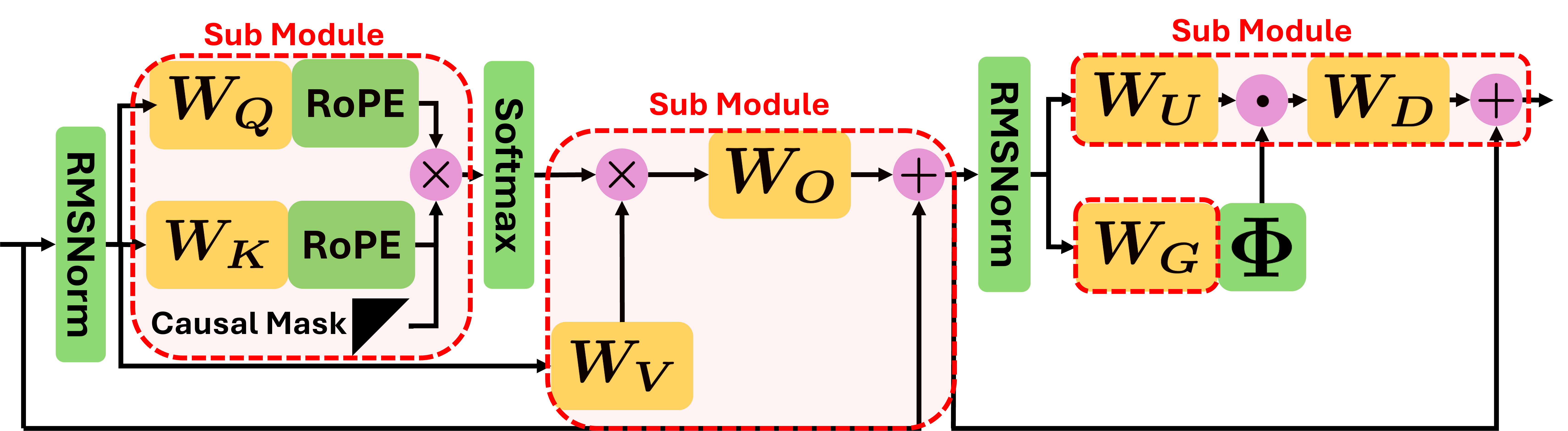}
    \caption{Conceptual diagram of the submodules considered in this work using LPCD; the regions enclosed by the red dashed boxes correspond to submodules.}
    \label{fig:submodule-LPCD}
\end{figure*}

\subsection{QEP as a Single LPCD Update}\label{subsec:qep-lpcd}
This section shows that a single iteration of LPCD corresponds to both QEP and LoaQ. 
For brevity, we explicitly demonstrate that QEP corresponds to one LPCD iteration; an analogous argument can be used to prove the LoaQ case.
To relate to QEP, we consider a two-block instance of the general formulation with the block variables $M_{1}=\widehat{W} \in \mab{R}^{N \times M}$ and $M_{2} = \widehat{X} \in \mab{R}^{T \times N}$.
The global objective is given by
\begin{equation}
    \label{eq:QEP-LPCD-objective}
    L(\widehat{W}, \widehat{X})=\|\widehat{X}\widehat{W}-XW\|_{F}^{2}.
\end{equation}
In this case, the following proposition holds:
\begin{proposition}
    Consider the objective defined in Eq.~\eqref{eq:QEP-LPCD-objective} with blocks $M_{1}=\widehat{W}$ and  $M_{2}=\widehat{X}$.
    Fix the activation block $\widehat{X}$ and perform a single LPCD update on the weight block $\widehat{W}$.
    Let $\widehat{W}^{(1)}$ denote the value of the weight block following this single LPCD update. Then $\widehat{W}^{\mathrm{QEP}} = \widehat{W}^{(1)}$.
\end{proposition}
\begin{proof}
The LPCD update of the weight block $M_{1}=\widehat{W}$ consists of two steps:

\paragraph{Relaxation Step.}
Fix $\widehat{X}$.
In the first outer iteration, the objective for $\widehat{W}$ takes the following form:
\begin{equation}
    \label{eq:qep-standard-least-square}
    L_{1}^{(1)}(U)=\|\widehat{X}^{(0)}U-XW\|_{F}^{2},
\end{equation}
where $\widehat{X}^{(0)}=\widehat{X}$.
Assume that $\widehat{H}$ is invertible. The optimality condition leads to the following minimizer:
\begin{equation}
    \overline{W}^{(1)}=\argmin_{U}L_{1}^{(1)}(U) = (I_{N} + \widehat{H}^{-1}C)W.
\end{equation}

\paragraph{Projection Step.}
For a weight block, LPCD employs a layer-wise PTQ projection using either the direct or activation-aware projection, $\Pi_{\mab{Q}}^{(d)}$ or $\Pi_{\mab{Q}}^{(a)}$, as defined in Section~\ref{sec:preliminaries}. 
In the QEP setting, the activation-aware projector $\Pi_{\mab{Q}}^{(a)}$ is employed. The projection step, therefore, reads 
\begin{equation}
    \label{eq:qep-lpcd-step2}
    \widehat{W}^{(1)}
    =
    \Pi_{\mab{Q}}^{(a)}\ab(\overline{W}^{(1)})
    =
    \Pi_{\mab{Q}}^{(a)}\ab((I_{N} + \widehat{H}^{-1} C) W).
\end{equation}
Finally, note that the QEP solution $\widehat{W}^{\mathrm{QEP}}$ from Eq.~\eqref{eq:qep-objective} satisfies
\begin{align}
    \widehat{W}^{\mathrm{QEP}} &= \argmin_{\widehat{W}} \|\widehat{X}\widehat{W} - X W\|_{F}^{2}  \\
    &= \argmin_{\widehat{W}} \|\widehat{X}(\widehat{W}-\overline{W}^{(1)})\|_{F}^{2}.
\end{align}
By definition of $\Pi_{\mab{Q}}^{(a)}$, the right-hand side coincides with the projection in Eq.~\eqref{eq:qep-lpcd-step2}. Therefore, $\widehat{W}^{\mathrm{QEP}} = \widehat{W}^{(1)}$ as stated.
\end{proof}

\begin{remark}
    An analogous argument demonstrates that LoaQ can also be interpreted as a single LPCD update for a suitably extended objective. For example, by considering
    \begin{equation}
        L(\widehat{W}, \widehat{X}, \widehat{R})= \|(\widehat{R} + \widehat{X} \widehat{W})-(R+XW)\|_{F}^{2},
    \end{equation}
    the same relaxation-projection decomposition yields $\widehat{W}^{\mathrm{LoaQ}} = \widehat{W}^{(1)}$, with $\widehat{W}^{(1)}$ computed by LPCD on this augmented submodule.
\end{remark}

Moreover, LPCD allows us to extend single-step algorithms that compensate for quantization error, such as QEP, to various settings.
Specifically, the extension of QEP to activation quantization is described in Appendix~\ref{subsec:QEP-activation}; its extension to KV-cache quantization is outlined in Appendix~\ref{subsec:qep-kv-cache}; and its extension to preprocessing using rotation matrices is detailed in Appendix~\ref{subsec:lpcd-rotation}.
All these algorithms can be implemented within a layer-wise PTQ framework, significantly reducing memory costs during quantization while minimizing quantization error.
Furthermore, both the QEP, viewed as a single-step method, and its extensions are expected to achieve higher performance by increasing the number of iterations in the alternating optimization process.

\begin{table*}[tb]
  \centering
  \caption{Perplexity (↓) on WikiText-2 for LLaMA and Qwen models across different bit-widths and quantization methods.}
  \label{tab:ppl_result}
  \setlength{\tabcolsep}{4pt}
  \begin{tabular}{c|cc|ccccc}
    \toprule
    \textbf{Bits} & \multicolumn{2}{c|}{\textbf{Method}} & 
    \textbf{LLaMA2-7B} & \textbf{LLaMA2-13B} & \textbf{LLaMA3-8B} & \textbf{Qwen3-8B} & \textbf{Qwen3-14B} \\
    \midrule
    FP16 & - & - & 4.8653 & 4.3560 & 5.4971 & 8.5980 & 7.5960 \\
      \midrule
      \multirow{6}{*}{4bit}
      & \multirow{3}{*}{RTN} & QEP & \textbf{5.2303} & 4.5432 & 6.9551 & 11.0644 & 12.7807 \\
      & & LoaQ & 5.3286 & \textbf{4.4831} & \textbf{6.1800} & 10.7548 & 8.5981 \\
      & & \qcol Ours & \qcol 5.2961 & \qcol 4.5237 & \qcol 7.4465 & \qcol \textbf{9.3566} & \qcol \textbf{8.1412} \\
      & \multirow{3}{*}{GPTQ} & QEP & 5.0954 & 4.4875 & 6.3459 & 10.9824 & 12.2558 \\
      & & LoaQ & 5.1399 & \textbf{4.4630} & \textbf{6.2109} & 9.9321 & 8.2968 \\
      & & \qcol Ours & \qcol \textbf{5.0495} & \qcol 4.4900 & \qcol 6.3818 & \qcol \textbf{9.1233} & \qcol \textbf{7.9668} \\
      \midrule
      \multirow{6}{*}{3bit}
      & \multirow{3}{*}{RTN} & QEP & 21.0619 & 6.3838 & 25.3924 & 22.5769 & 18.1640 \\
      & & LoaQ & 8.9140 & 5.4920 & 14.1467 & >1e3 & 12.1034 \\
      & & \qcol Ours & \qcol \textbf{6.5760} & \qcol \textbf{5.3979} & \qcol \textbf{9.8112} & \qcol \textbf{12.7110} & \qcol \textbf{10.8723} \\
      & \multirow{3}{*}{GPTQ} & QEP & 6.3966 & 5.1518 & 11.0124 & 15.0779 & 14.2997 \\
      & & LoaQ & 6.8494 & \textbf{4.9777} & 9.0706 & 11.7249 & 10.8154 \\
      & & \qcol Ours & \qcol \textbf{5.8990} & \qcol 5.0785 & \qcol \textbf{8.7971} & \qcol \textbf{11.3805} & \qcol \textbf{9.5815} \\
      \midrule
      \multirow{6}{*}{2bit}
      & \multirow{3}{*}{RTN} & QEP & >1e3 & >1e3 & >1e3 & >1e3 & >1e3 \\
      & & LoaQ & >1e3 & >1e3 & >1e3 & 375.1837 & 831.7296 \\
      & & \qcol Ours & \qcol >1e3 & \qcol \textbf{552.2888} & \qcol >1e3 & \qcol \textbf{312.4608} & \qcol \textbf{352.2354} \\
      & \multirow{3}{*}{GPTQ} & QEP & \textbf{101.1521} & 84.3543 & >1e3 & 165.7484 & 199.1968 \\
      & & LoaQ & 590.9850 & \textbf{24.2423} & 217.8416 & 550.3343 & \textbf{43.2624} \\
      & & \qcol Ours & \qcol 341.3434 & \qcol 26.3311 & \qcol \textbf{87.5296} & \qcol \textbf{58.8030} & \qcol 46.4656 \\
    \bottomrule
  \end{tabular}
\end{table*}

\subsection{Submodule PTQ}\label{subsec:submodule-wise-ptq}

In this study, we apply LPCD to several submodules for which the relaxation step provides a closed-form solution. Although closed-form expressions exist, some of these problems are memory-inefficient to solve exactly; therefore, so we approximate the relaxation step in practice; see Appendix \ref{subsec:update-submodule}.
A conceptual diagram of this procedure is presented in Figure~\ref{fig:submodule-LPCD}.
In the following, we briefly describe how the proposed method is applied to each submodule.
Note that the proposed method can also be applied to submodules with nonlinear transformations by approximating the minimization; a more exhaustive study of such applications is left for future work.

\paragraph{QK Module.}
We consider a setting in which grouped-query attention is quantized at the level of its QK submodule.
Specifically, a single key and value are shared within each group of $G$ heads.
The objective of the QK module is expressed as
\begin{multline}
    \label{eq:QK-module-objective}
    L(\hat{W}_{Q}, \hat{W}_{K}) = 
    \sum_{h \in [H]} \|M \odot (\widehat{S}^{(h)}-S^{(h)}) \|_{F}^{2} \\
    \widehat{S}^{(h)} = \mac{R}(\widehat{X} \widehat{W}_{Q}^{(h)})\mac{R}(\widehat{X} \widehat{W}_{K}^{(g)})^{\top},\\
    S^{(h)} = R(X W_{Q}^{(h)})R(X W_{K}^{(g)})^{\top}.
\end{multline}
where $g = \lfloor \nicefrac{h-1}{G} \rfloor$ and $M \in \{0, 1\}^{T \times T}$ are binary upper-triangular matrices that represent the causal mask.
The operator $\mac{R}(\cdot)$ denotes rotary positional encoding (RoPE).
When either $\widehat{W}_{Q}$ or $\widehat{W}_{K}$ is fixed, the objective reduces to a linear least-squares formulation, analogous to QEP.
The detailed update rule for the relaxation step is provided in Appendix \ref{subsub:kq-module}

\paragraph{VO Module.}
Next, we consider the submodule that aggregates the attention scores $S^{(h)} = \mathrm{Softmax}(S^{(h)})$ obtained by applying the softmax function after the KV module, and we quantize this component.
Specifically, the objective can be expressed as follows:
\begin{multline}
    L(\widehat{W}_{V}, \widehat{W}_{O}) = \|\widehat{\Omega} + \widehat{R} - (\Omega+R)\|_{F}^{2}, \\
    \widehat{\Omega} = \mathrm{Concat}_{h \in [H]}(\widehat{S}^{(h)} (\widehat{X} \widehat{W}_{V}^{(g)})) \widehat{W}_{O}, \\
    \Omega = \mathrm{Concat}_{h \in [H]}(S^{(h)}(X W_{V}^{(g)})) W_{O},
\end{multline}
where $\mathrm{Concat}_{h \in [H]}$ denotes concatenation along the head dimension. As in the previous case, the minimization becomes straightforward once either $\widehat{W}_{V}$ or $\widehat{W}_{O}$ is fixed.
Further details are provided in  Appendix \ref{subsec:up-down-module}.

\paragraph{Up-Down Module.}
After the self-attention block, most Transformer architectures process the representations through an MLP layer. We quantize the Up–Down projection in this MLP as a submodule. The objective function is expressed as
\begin{multline}
    L(\widehat{W}_{U}, \widehat{W}_{D}) = \|\widehat{F}+\widehat{R} - (F+R) \|_{F}^{2},\\
    \widehat{F} = \ab(\Phi(\widehat{X}\widehat{W}_{G}) \odot \widehat{X}\widehat{W}_{U}) \widehat{W}_{D},\\
    F = \ab(\Phi(XW_{G}) \odot XW_{U}) W_{D},
\end{multline}
where $\Phi(\cdot)$ denotes the activation function, and LLaMA employs the SiLU function \citep{touvron2023llama}.
This work restricts the optimization variables to $W_{U}$ and $W_{D}$ to simplify the minimization process in the relaxation Step.
However, LPCD can also be applied by approximately solving the minimization problem concerning $W_{G}$; investigating the effect of this approach is left for future work.
The detailed update rules for this Up-Down submodule are provided in Appendix \ref{subsec:up-down-module}.

\section{Experiments}

\begin{table*}[tb]
  \centering
  \caption{Zero-shot average accuracy (↑) on ARC-E and PIQA for LLaMA and Qwen models across different bit-widths and quantization methods.}
  \label{tab:acc_result}
  \setlength{\tabcolsep}{4pt}
  \begin{tabular}{c|cc|ccccc}
    \toprule
    \textbf{Bits} & \multicolumn{2}{c|}{\textbf{Method}} & 
    \textbf{LLaMA2-7B} & \textbf{LLaMA2-13B} & \textbf{LLaMA3-8B} & \textbf{Qwen3-8B} & \textbf{Qwen3-14B} \\
    \midrule
    FP16 & - & - & 0.7682 & 0.7896 & 0.7915 & 0.7930 & 0.8133 \\
      \midrule
      % ========== 4bit ==========
      \multirow{6}{*}{4bit}
      & \multirow{3}{*}{RTN} & QEP & 0.7467 & \textbf{0.7825} & 0.7613 & 0.6998 & 0.7369 \\
      & & LoaQ & 0.7471 & 0.7812 & \textbf{0.7812} & 0.6333 & 0.7760 \\
      & & \qcol Ours & \qcol \textbf{0.7527} & \qcol 0.7793 & \qcol 0.7789 & \qcol \textbf{0.7584} & \qcol \textbf{0.7949} \\
      & \multirow{3}{*}{GPTQ} & QEP & 0.7428 & 0.7741 & \textbf{0.7599} & 0.7113 & 0.7695 \\
      & & LoaQ & 0.7456 & \textbf{0.7817} & 0.7559 & 0.6066 & 0.7702 \\
      & & \qcol Ours & \qcol \textbf{0.7504} & \qcol 0.7810 & \qcol 0.7001 & \qcol \textbf{0.7661} & \qcol \textbf{0.7782} \\
      \midrule
      % ========== 3bit ==========
      \multirow{6}{*}{3bit}
      & \multirow{3}{*}{RTN} & QEP & \textbf{0.5784} & 0.6757 & 0.5136 & 0.5442 & 0.6164 \\
      & & LoaQ & 0.4799 & 0.6928 & 0.5282 & 0.3971 & \textbf{0.6904} \\
      & & \qcol Ours & \qcol 0.5688 & \qcol \textbf{0.7433} & \qcol \textbf{0.5373} & \qcol \textbf{0.6291} & \qcol 0.6560 \\
      & \multirow{3}{*}{GPTQ} & QEP & \textbf{0.6881} & 0.7372 & 0.5427 & 0.5986 & 0.6535 \\
      & & LoaQ & 0.4556 & \textbf{0.7462} & 0.5532 & 0.6363 & 0.6222 \\
      & & \qcol Ours & \qcol 0.6602 & \qcol 0.7431 & \qcol \textbf{0.6290} & \qcol \textbf{0.6493} & \qcol \textbf{0.6919} \\
      \midrule
      % ========== 2bit ==========
      \multirow{6}{*}{2bit}
      & \multirow{3}{*}{RTN} & QEP & \textbf{0.3816} & \textbf{0.3831} & \textbf{0.3904} & 0.3845 & 0.3895 \\
      & & LoaQ & 0.3758 & 0.3805 & 0.3836 & \textbf{0.3917} & \textbf{0.4183} \\
      & & \qcol Ours & \qcol 0.3756 & \qcol 0.3794 & \qcol 0.3826 & \qcol 0.3864 & \qcol 0.3864 \\
      & \multirow{3}{*}{GPTQ} & QEP & \textbf{0.3931} & 0.3821 & 0.3889 & 0.3972 & \textbf{0.4355} \\
      & & LoaQ & 0.3761 & \textbf{0.3840} & 0.3842 & 0.4031 & 0.4253 \\
      & & \qcol Ours & \qcol 0.3856 & \qcol 0.3817 & \qcol \textbf{0.3962} & \qcol \textbf{0.4054} & \qcol 0.4354 \\
    \bottomrule
  \end{tabular}
\end{table*}

We conduct experiments to assess the effectiveness of the proposed method.
To evaluate how much LPCD enhances the existing approach,  we utilize the final weight of QEP or LoaQ and combine LPCD with the current quantization method.

\subsection{Setting}\label{subsec:setting}

\paragraph{Baselines and Quantization Methods.}
In this study, we focus only on the per-channel weight quantization scheme.
We employ representative layer-wise PTQ methods of Round-to-nearest (RTN) and GPTQ, which are used in conjunction with error compensation techniques.
Some error compensation methods have been reported to enhance existing layer-wise quantization methods. We employ QEP and LoaQ as baselines for our proposed method.
We perform quantization in the INT4, INT3, and INT2 settings. 
We skip quantization for the last 2 layers due to the higher frequency of outliers observed in their activations.

\paragraph{Dataset.}
GPTQ computes Hessian using the calibration dataset to perform effective quantization.
Furthermore, error compensation methods compute activations for the output approximation.
Following previous studies, LoaQ uses 128 samples of 2048 tokens each from the C4 dataset.
We observe over-fitting to the calibration dataset when we employ 128 samples.
We employ 2048 tokens consisting of 256 sequences, randomly sampled from the WikiText-2 dataset, as the calibration dataset.

\paragraph{Models.}
We evaluate the proposed method and baselines in recent major open-weight LLM, including LLaMA2~\citep{touvron2023llama}, LLaMA3~\citep{grattafiori2024llama}, and the dense Qwen3 model families.
LLaMA is an open-weight LLM family that is primarily developed by Meta Platforms.  
We employ LLaMA2-7B, LLaMA2-13B, and LLaMA3-8B for evaluation.
Qwen3 is a powerful open-weight LLM family as well as LLaMA family.
They employ a slightly different architecture from LLaMA such as Q/K RMSNorm. We employ Qwen3-4B, Qwen3-8B, and Qwen3-14B for evaluation.

\paragraph{Hyper Parameters.}
We conduct a grid search to determine the optimal propagation strength parameter $\alpha$ for QEP and the optimal sub-layer output approximation strength parameter $\beta$ for LoaQ.
The grid search is performed using smaller models, such as Qwen3-0.5B and LLaMA3.2-1B, after which we applied the resulting optimal parameters to the larger models.
Following LoaQ, the range of $\alpha$ is from $0$ to $1$ with increments of 0.1 and, the range of $\beta$ is from $0$ to $1$ with increments 0.05.
We apply LPCD on the relaxed weight of LoaQ; performances of LoaQ are generally better than those of QEP.
We first apply LoaQ to each submodule and then perform LPCD.
As explained in Sec.~\ref{subsec:submodule-wise-ptq}, we apply LPCD to three groups of Transformer submodules: the Q/K module, the V/O module, and the Up/Down module.
For gradient-based optimization of LPCD, We employ 8 batch size, 40 epochs, and cosine scheduled learning rate that begin with $10^{-5}$.
The optimization is conducted using the Adam optimizer with the default settings in \texttt{PyTorch}.

\paragraph{Evaluations.}
We follow the established evaluation protocols for quantization algorithms used in numerous previous studies.
We evaluate the perplexity (PPL) on the WikiText2 dataset. We also evaluated zero-shot accuracy on the ARC Easy and PiQA.benchmarks. 
We implement QEP, LoaQ, and LPCD using Python 3.12.11 with PyTorch 2.4.0 and Hugging Face Transformers 4.55.3.
All experiments were conducted on an NVIDIA H100 GPU using the TSUBAME 4.0 supercomputer.

\subsection{Result}

\paragraph{Perplexities.}
Table~\ref{tab:ppl_result} summarizes the perplexities of various PTQ configurations within the LLaMA and Qwen families. Overall, LPCD-based submodule quantization achieves the lowest perplexity in most settings, consistently outperforming both QEP and LoaQ, irrespective of whether RTN or GPTQ is used. 
The gains are most pronounced in low-bit regimes: for the practically important LLaMA-3-8B and Qwen-3-8B models at 3-bit and 2-bit levels, LPCD substantially reduces PPL compared to both baselines, preventing the severe degradation or divergence observed with QEP and LoaQ. Notably, for Qwen-3-8B, RTN combined with LPCD already surpasses the more sophisticated QEP+GPTQ configuration, indicating that submodule LPCD provides improvements that are largely orthogonal to the choice of the underlying layer-wise quantizer.

\paragraph{Zero Shot Task Evaluation.}

Table 2 indicates that LPCD achieves the highest or nearly the highest zero-shot accuracy across various models, bit-widths, and base quantizers. At 4-bits, our method closely matches FP16 performance while slightly improving both QEP and LoaQ, indicating that optimization at the submodule-level does not adversely affect high-precision behavior.
The advantages are more apparent in low-bit regimes. For the practically important LLaMA-3-8B and Qwen-3-8B models at 3-bit and 2-bit, LPCD consistently recovers a substantial portion of the accuracy lost by QEP and LoaQ when used with RTN and GPTQ. Remarkably, the simple RTN+LPCD configuration on Qwen-3-8B outperforms the more sophisticated QEP+GPTQ baseline, demonstrating that our submodule refinement complements rather than merely imitates existing layer-wise PTQ techniques.

\section{Conclusion}\label{sec:conclusion}

We propose LPCD, a unified framework that extends PTQ beyond traditional layer-wise formulations. 
LPCD optimizes relaxed objectives across arbitrary Transformer submodules and subsequently projects the solutions back using existing layer-wise PTQ projectors. 
In this formulation, GPTQ-style activation-aware PTQ, QEP, and LoaQ are presented as special cases, each corresponding to specific choices of submodules, loss functions, and single-step updates. 
Thus, LPCD can be viewed as a strict generalization that unifies previously separate approaches into a submodule-centric perspective while preserving the efficiency and modularity of standard layer-wise pipelines.

For the LLaMA and Qwen models, LPCD-based submodule quantization results in lower perplexity and shows competitive or superior zero-shot accuracy compared to QEP and LoaQ across various bit-widths, particularly at 3 and 2 bits, without altering the underlying layer-wise quantizers or the inference stack. 
These results indicate that optimizing at the submodule level with LPCD yields consistent gains in addition to existing QEP and LoaQ-style refinements, rather than replacing them. 
Future work includes applying LPCD to more complex nonlinear submodules, jointly handling weights, activations, and KV caches at scale, and integrating LPCD with quantization-aware finetuning to further enhance the deployability of low-bit LLMs.

\section*{Acknowledgements}
The authors would like to express their sincere gratitude to Koichi Shirahata and Yuhei Umeda of Fujitsu Limited, and to Katsuki Fujisawa, Toshio Endo, and Yoshihiko Fujisawa of Institute of Science Tokyo for their valuable support and insightful advice.
This work was partially supported by JST BOOST, Japan (Grant No. JPMJBY24D0), and by the Cabinet Office, Government of Japan, through the SIP program ``Promotion of the Application of Advanced Quantum Technology Platforms to Social Issues''.

\bibliography{ref}

\newpage

\appendix
\onecolumn

\section{Derivation}\label{app-sec:derivation}

\subsection{Update Rule of Submodule PTQ}\label{subsec:update-submodule}
In this section, we derive the relaxation step for the QK, VO, and Up-Down modules introduced in Sec. \ref{subsec:submodule-wise-ptq}.
In all cases, the objective reduces to a linear least-squares problem once one block is fixed, allowing the relaxed minimizer to have a closed-form expression. However, the associated design matrices are exceptionally large in realistic LLM configurations, rendering the exact closed-form solution memory-inefficient. Therefore, we solve the least-squares problems approximately using gradient-based methods

\subsubsection{QK Module}\label{subsub:kq-module}

We consider a single head of grouped-query attention; the multi-head and grouped cases are derived by applying the same derivation either head-wise or group-wise.
Let $S_{M} \in \mab{R}^{T \times T}$ denote the full-precision causal-masked attention score matrix for this head, i.e., $S_{M} = M \odot S$ for some unmasked score matrix $S$ and a fixed causal mask $M \in \{0,1\}^{T\times T}$.
Let
$\widehat{W}_{Q}, \widehat{W}_{K} \in \mab{R}^{D_{\mathrm{model}} \times d_{k}}$ be the quantized query and key projection matrices, where $d_{k}$ denotes the head dimension.
The RoPE operator $\mac{R}(\cdot)$ is linear in its argument.
Therefore, there exists a fixed block-diagonal matrix
$R_{\Theta} \in \mab{R}^{T d_{k} \times T d_{k}}$ such that for any
$Z \in \mab{R}^{T \times d_{k}}$, 
$\operatorname{vec}\ab(\mac{R}(Z)) = R_{\Theta}\operatorname{vec}(Z).$
The QK-submodule loss for this head is expressed as 
\begin{equation}
    L(\widehat{W}_{Q}, \widehat{W}_{K}) =
    \ab\|M \odot \ab(\mac{R}(\widehat{X}\widehat{W}_{Q})\mac{R}(\widehat{X}\widehat{W}_{K})^{\top}- S_{M})\|_{F}^{2}.
    \label{eq:qk-objective-single-head}
\end{equation}
Using $\|A\|_{F}^{2} = \|\operatorname{vec}(A)\|_{2}^{2}$ along with
$\operatorname{vec}(AB^{\top}) = (B \otimes I_{T})\operatorname{vec}(A)$ and
$\operatorname{vec}(\widehat{X}\widehat{W}_{Q}) = (I_{d_{k}} \otimes \widehat{X})\operatorname{vec}(\widehat{W}_{Q})$,
we can express Eq.~
\eqref{eq:qk-objective-single-head} as
\begin{equation}
    L(\widehat{W}_{Q}, \widehat{W}_{K})
    = \ab\|D_{M} Z_{Q} \operatorname{vec}(\widehat{W}_{Q})- \operatorname{vec}(S_{M})\|_{2}^{2},
\end{equation}
where $D_{M} = \operatorname{diag}(\operatorname{vec}(M)) \in \mab{R}^{T^{2}\times T^{2}}$ and
\begin{equation}
    Z_{Q} = \ab(\mac{R}(\widehat{X}\widehat{W}_{K}) \otimes I_{T})
    R_{\Theta}(I_{d_{k}} \otimes \widehat{X})
    \in \mab{R}^{T^{2} \times (D_{\mathrm{model}} d_{k})}.
\end{equation}
An analogous expression is obtained when $\widehat{W}_{Q}$ is fixed and $L$ is considered a function of $\widehat{W}_{K}$.

\paragraph{Query Relaxation Step.}
Fix $\widehat{W}_{K}$.
Then $Z_{Q}$ is constant, and the loss represents a quadratic function of
$u = \operatorname{vec}(\widehat{W}_{Q})$.
\begin{equation}
    \operatorname{vec}(\overline{W}_{Q}^{(1)}) = \argmin_{u}
    \ab\|D_{M} Z_{Q} u - \operatorname{vec}(S_{M})\|_{2}^{2}.
    \label{eq:qk-query-relaxed}
\end{equation}
This is a linear least-squares problem.
When $Z_{Q}^{\top} D_{M}^{\top} D_{M} Z_{Q}$ is invertible, the unique minimizer is
\begin{equation}
    \operatorname{vec}(\overline{W}_{Q}^{(1)})
    = \ab(Z_{Q}^{\top} D_{M}^{\top} D_{M} Z_{Q})^{-1}
    Z_{Q}^{\top} D_{M}^{\top} \operatorname{vec}(S_{M}),
\end{equation}
and, more generally,
\begin{equation}
    \operatorname{vec}(\overline{W}_{Q}^{(1)}) =
    (D_{M} Z_{Q})^{\dagger} \operatorname{vec}(S_{M}),
\end{equation}
where $(\cdot)^{\dagger}$ denotes the Moore–Penrose pseudoinverse.

\paragraph{Key Relaxation Step.}
Fix $\widehat{W}_{Q}$. By symmetry, we define
\begin{equation}
    Z_{K} = \ab(\mac{R}(\widehat{X}\widehat{W}_{Q}) \otimes I_{T})
    R_{\Theta} \ab(I_{d_{k}} \otimes \widehat{X})
    \in \mab{R}^{T^{2} \times (D_{\mathrm{model}} d_{k})},
\end{equation}
so that
\begin{equation}
    L(\widehat{W}_{Q}, \widehat{W}_{K}) =\ab\|
        D_{M} Z_{K} \operatorname{vec}(\widehat{W}_{K}) - \operatorname{vec}(S_{M}) \|_{2}^{2}.
\end{equation}
The key relaxation step is therefore:
\begin{equation}
    \operatorname{vec}(\overline{W}_{K}^{(1)})
    = \argmin_{u}
    \ab\|D_{M} Z_{K} u - \operatorname{vec}(S_{M})\|_{2}^{2},
    \label{eq:qk-key-relaxed}
\end{equation}
which is again a linear least-squares problem.
When $Z_{K}^{\top} D_{M}^{\top} D_{M} Z_{K}$ is invertible, a unique minimizer exists.
\begin{equation}
    \operatorname{vec}(\overline{W}_{K}^{(1)})
    = \ab(Z_{K}^\top D_{M}^\top D_{M} Z_{K})^{-1}
    Z_{K}^\top D_{M}^\top \operatorname{vec}(S_{M}),
\end{equation}
In general,
\begin{equation}
    \operatorname{vec}(\overline{W}_{K}^{(1)})
    =
    (D_{M} Z_{K})^{\dagger} \operatorname{vec}(S_{M}).
\end{equation}

\paragraph{Memory Cost.}
For each head, the design matrices $Z_{Q}$ and $Z_{K}$ are of size
$T^{2} \times (D_{\mathrm{model}} d_{k})$.
Storing $Z_{Q}$ or $Z_{K}$ explicitly requires $\Theta(T^{2} D_{\mathrm{model}} d_{k})$ memory, while forming the normal matrices $Z_{Q}^\top D_{M}^\top D_{M} Z_{Q}$ and
$Z_{K}^\top D_{M}^\top D_{M} Z_{K}$ requires $\Theta(T^{2} (D_{\mathrm{model}} d_{k})^{2})$ operations. For typical LLM configurations, this is prohibitive.
In our implementation, we therefore do not employ $Z_{Q}$ or $Z_{K}$ and do not compute the closed-form solutions mentioned above; instead, we approximate the minimizers of Eq.~
\eqref{eq:qk-query-relaxed} and Eq.~ \eqref{eq:qk-key-relaxed} using gradient-based least-squares solvers applied directly to the original objective, Eq.~\eqref{eq:qk-objective-single-head}.

\subsubsection{VO Module}\label{subsec:vo-module}

We consider the VO submodule.
Let
\begin{equation}
    Y = \Omega + R,~~~\widehat{Y} = \widehat{\Omega} + \widehat{R},
\end{equation}
where $R$ and $\widehat{R}$ denote the full-precision and quantized residual streams, respectively, and
\begin{align}
    \Omega = \mathrm{Concat}_{h \in [H]}
    \ab(S^{(h)} V^{(g(h))}) W_{O},~~\widehat{\Omega} =
    \mathrm{Concat}_{h \in [H]}\ab(\widehat{S}^{(h)} \widehat{V}^{(g(h))}) \widehat{W}_{O}.
\end{align}
Here, $g(h)$ is the group index of head $h$;
$V^{(g)} = X W_{V}^{(g)}$ and $\widehat{V}^{(g)} = \widehat{X}\widehat{W}_{V}^{(g)}$ are the value projections for group $g$, and
$S^{(h)}$ and $\widehat{S}^{(h)}$ are the full-precision and quantized attention weights for head $h$. The VO submodule loss can be expressed as 
\begin{equation}
    L(\widehat{W}_{V}, \widehat{W}_{O})
    = \ab\|\widehat{Y} - Y\|_{F}^{2}.
\end{equation}

\paragraph{Value Relaxation Step.}
Fix $\widehat{W}_{O}$ and all other blocks.
For a fixed group $g$, only the value weights $\widehat{W}_{V}^{(g)}$ are updated, while all
other quantities are treated as constants.
Let $\mathcal{H}_{g}$ be the set of heads in group $g$ and let $d_{v}$ be the per-head value
dimension.
We decompose the quantized output as
\begin{equation}
\widehat{Y}
=
\widehat{Y}_{\neg g} + \widehat{Y}_{g},
~~~
\widehat{Y}_{g}
=
\sum_{h \in \mathcal{H}_{g}}
  \widehat{S}^{(h)} \widehat{X}\widehat{W}_{V}^{(g)} \widehat{W}_{O}^{(h)},
\end{equation}
where $\widehat{Y}_{\neg g}$ collects all contributions that do not involve
$\widehat{W}_{V}^{(g)}$, and $\widehat{W}_{O}^{(h)} \in \mathbb{R}^{d_{v}\times D_{\mathrm{model}}}$ represents
the output projection for head $h$.
The blockwise objective can be written as
\begin{equation}
\min_{\widehat{W}_{V}^{(g)}}
\ab\|
  Y - \widehat{Y}_{\neg g} - \widehat{Y}_{g}
\|_{F}^{2}.
\end{equation}

By stacking the heads and applying standard vectorization identities, this objective is
equivalently formulated as follows:
\begin{equation}
\min_{u}
\ab\|
  Z_{V}^{(g)} u - y^{\ast}
\|_{2}^{2},
~~~
u = \operatorname{vec}\ab(\widehat{W}_{V}^{(g)}),\quad
y^{\ast} = \operatorname{vec}\ab(Y - \widehat{Y}_{\neg g}),
\end{equation}
where
\begin{equation}
    Z_{V}^{(g)}
    =
    \sum_{h \in \mathcal{H}_{g}}
      \ab(\widehat{W}_{O}^{(h)\top} \otimes \widehat{S}^{(h)} \widehat{X})
    \in \mathbb{R}^{T D_{\mathrm{model}} \times D_{\mathrm{model}} d_{v}}.
\end{equation}
Thus, the relaxed minimizer for $\widehat{W}_{V}^{(g)}$ is the least-squares solution
\begin{equation}
    \operatorname{vec}(\overline{W}_{V}^{g,(1)})
    =
    \argmin_{u}
    \ab\|
        Z_{V}^{(g)} u - y^{\ast}
    \|_{2}^{2}.
\end{equation}
If $\ab(Z_{V}^{(g)})^{\top} Z_{V}^{(g)}$ is invertible, the unique minimizer is
\begin{equation}
    \operatorname{vec}(\overline{W}_{V}^{g,(1)})
    =
    \ab(
        \ab(Z_{V}^{(g)})^\top Z_{V}^{(g)}
    )^{-1}
    \ab(Z_{V}^{(g)})^\top y^{\ast},
    \label{eq:vo-value-relaxed}
\end{equation}
We may replace the inverse with the Moore–Penrose pseudoinverse.

\paragraph{Output-Projection Relaxation Step.}
Fix $\widehat{W}_{V}$ and all other blocks.
Let
\begin{equation}
    \widehat{H}
    =
    \mathrm{Concat}_{h \in [H]}
    \ab(
        \widehat{S}^{(h)} \widehat{V}^{(g(h))}
    )
    \in \mathbb{R}^{T \times (H d_{v})},
\end{equation}
where $\widehat{V}^{(g(h))} = \widehat{X}\widehat{W}_{V}^{(g(h))}$ represents the value projection for the
group that includes head $h$.
Then, the attention output can be expressed as $\widehat{\Omega} = \widehat{H} \widehat{W}_{O}$, and the blockwise objective for output projection is
\begin{equation}
    \min_{\widehat{W}_{O}}
    \ab\|
        \widehat{H} \widehat{W}_{O} - (Y - \widehat{R})
    \|_{F}^{2}.
\end{equation}
This is a standard least-squares matrix problem.
If $\widehat{H}^{\top}\widehat{H}$ is invertible, the relaxed minimizer is expressed in closed form as follows:
\begin{equation}
    \overline{W}_{O}^{(1)}
    =
    \ab(\widehat{H}^{\top}\widehat{H})^{-1}
    \widehat{H}^{\top}\,(Y - \widehat{R}),
    \label{eq:vo-out-relaxed}
\end{equation}
and, more generally
$\overline{W}_{O}^{\ast}
 = (\widehat{H}^{\top}\widehat{H})^{\dagger}\widehat{H}^{\top}(Y - \widehat{R})$
when $\widehat{H}^{\top}\widehat{H}$ is not invertible.

\paragraph{Memory Cost.}
For each group $g$, the design matrix $Z_{V}^{(g)}$ is of size
$T D_{\mathrm{model}} \times (D_{\mathrm{model}} d_{v})$, making the explicit formation of $Z_{V}^{(g)}$ and the
computation of $(Z_{V}^{(g)})^{\top} Z_{V}^{(g)}$ memory- and compute-intensive at LLM scale.
In practice, we do not materialize $Z_{V}^{(g)}$; instead, we approximately solve the
least-squares problem in Eq.~\eqref{eq:vo-value-relaxed} directly using a gradient-based method on the
original VO loss.
In contrast, the output-projection relaxation step involves only 
$\widehat{H} \in \mathbb{R}^{T \times (H d_{v})}$ and
$\widehat{H}^{\top}\widehat{H} \in \mathbb{R}^{(H d_{v}) \times (H d_{v})}$,
whose sizes depend on the number of heads and the value dimension per-head, but not on
$D_{\mathrm{model}}$.
Thus, the closed-form solution in Eq.~\eqref{eq:vo-out-relaxed} is computationally feasible, allowing us to compute
the O-step exactly without relying on gradient-based optimization.

\subsubsection{Up–Down Module}\label{subsec:up-down-module}

Finally, we consider the Up–Down MLP submodule.
Let
\begin{equation}
    \Phi = \Phi(X W_{G}), ~~~
    \widehat{\Phi} = \Phi(\widehat{X} \widehat{W}_{G}),~~~F
    =
    \ab(
        \Phi \odot X W_{U}
    ) W_{D},
    ~~~
    \widehat{F}
    =
    \ab(
        \widehat{\Phi} \odot \widehat{X} \widehat{W}_{U}
    ) \widehat{W}_{D},
\end{equation}
where $\odot$ denotes the element-wise multiplication.
The submodule loss is
\begin{equation}
    L(\widehat{W}_{U}, \widehat{W}_{D})
    =
    \ab\|
        \widehat{F} + \widehat{R} - (F + R)
    \|_{F}^{2}
    =
    \ab\|
        \widehat{F} - Y_{\mathrm{MLP}}
    \|_{F}^{2},
\end{equation}
with $Y_{\mathrm{MLP}} = F + R - \widehat{R}$.

\paragraph{Up Relaxation Step.}
Fix $\widehat{W}_{D}$ and all other blocks, and set
\begin{equation}
    \widehat{U}
    =
    \widehat{X} \widehat{W}_{U}
    \in \mathbb{R}^{T \times D_{\mathrm{up}}},
    ~~
    \widehat{\Phi}
    =
    \Phi(\widehat{X} \widehat{W}_{G})
    \in \mathbb{R}^{T \times D_{\mathrm{up}}},~~\widehat{F}
    =
    \ab(\widehat{\Phi} \odot \widehat{U}) \widehat{W}_{D}.
\end{equation}
Let $D_{\Phi} = \operatorname{diag}(\operatorname{vec}(\widehat{\Phi}))$.
Using standard vectorization identities, we obtain
\begin{equation}
    \operatorname{vec}(\widehat{F})
    =
    Z_{U}\, \operatorname{vec}(\widehat{W}_{U}),
    ~~~
    Z_{U}
    =
    \ab(\widehat{W}_{D}^\top \otimes I_{T})
    D_{\Phi}
    \ab(I_{D_{\mathrm{up}}} \otimes \widehat{X})
    \in
    \mathbb{R}^{T D_{\mathrm{model}} \times D_{\mathrm{model}} D_{\mathrm{up}}}.
\end{equation}
Thus, the relaxation step for $\widehat{W}_{U}$ reduces to a linear least-squares problem
\begin{equation}
    \operatorname{vec}(\overline{W}_{U}^{(1)})
    =
    \argmin_{u}
    \ab\|
        Z_{U} u - \operatorname{vec}(Y_{\mathrm{MLP}})
    \|_{2}^{2}.
    \label{eq:updown-up-relaxed}
\end{equation}
In principle, when $Z_{U}^\top Z_{U}$ is invertible, the unique minimizer is given by
\begin{equation}    \operatorname{vec}(\overline{W}_{U}^{(1)})
    =
    \ab(Z_{U}^\top Z_{U})^{-1}
    Z_{U}^\top \operatorname{vec}(Y_{\mathrm{MLP}}),
\end{equation}
or more generally by $Z_{U}^{\dagger} \operatorname{vec}(Y_{\mathrm{MLP}})$ using the Moore–Penrose
pseudoinverse.
However, $Z_{U}$ has a size
of $T D_{\mathrm{model}} \times (D_{\mathrm{model}} D_{\mathrm{up}})$ and is prohibitively large to be
formed explicitly for LLM-scale models.
In practice, we approximate the solution of Eq.~\eqref{eq:updown-up-relaxed} using a
gradient-based least-squares solver directly on the loss
$\|\widehat{F} - Y_{\mathrm{MLP}}\|_{F}^{2}$, without explicitly constructing $Z_{U}$.

\paragraph{Down Relaxation Step.}
Fix $\widehat{W}_{U}$ and all other blocks.
Recall that
\begin{equation}
    \widehat{F}
    =
    \ab(\widehat{\Phi} \odot \widehat{X}\widehat{W}_{U})\widehat{W}_{D}.
\end{equation}
Define
\begin{equation}
    \widehat{Z}_{D}
    \coloneqq
    \widehat{\Phi} \odot \widehat{X}\widehat{W}_{U}
    \in \mathbb{R}^{T \times D_{\mathrm{up}}},~~\widehat{F} = \widehat{Z}_{D}\widehat{W}_{D}.
\end{equation}
Substituting this into the submodule loss yields the blockwise objective
\begin{equation}
    \min_{\widehat{W}_{D}}
    \ab\|
        \widehat{Z}_{D}\widehat{W}_{D} - Y_{\mathrm{MLP}}
    \|_{F}^{2}.
\end{equation}
This is a standard matrix least-squares problem concerning 
$\widehat{W}_{D} \in \mathbb{R}^{D_{\mathrm{up}}\times D_{\mathrm{model}}}$.
If $\widehat{Z}_{D}^{\top}\widehat{Z}_{D}$ is invertible, then the unique minimizer is
\begin{equation}
    \overline{W}_{D}^{(1)}
    =
    \ab(\widehat{Z}_{D}^{\top}\widehat{Z}_{D})^{-1}
    \widehat{Z}_{D}^{\top} Y_{\mathrm{MLP}},
    \label{eq:updown-down-relaxed}
\end{equation}
and, more generally, $\overline{W}_{D}^{(1)} = \widehat{Z}_{D}^{\dagger} Y_{\mathrm{MLP}}$ when $\widehat{Z}_{D}^{\top}\widehat{Z}_{D}$ is not invertible, where
$\widehat{Z}_{D}^{\dagger}$ denotes the Moore–Penrose pseudoinverse of $\widehat{Z}_{D}$.

\paragraph{Memory Cost.}
The design matrix $Z_{U}$ has a size of $T D_{\mathrm{model}} \times (D_{\mathrm{model}} D_{\mathrm{up}})$, and constructing the normal matrix $Z_{U}^\top Z_{U}$ requires $\Theta(T D_{\mathrm{model}}^{2} D_{\mathrm{up}}^{2})$ operations, which is prohibitive at the scale of LLM.
We therefore do not materialize $Z_{U}$ and instead solve the least-squares problem in Eq.~\eqref{eq:updown-up-relaxed} approximately by applying a gradient-based method directly to $\|\widehat{F} - Y_{\mathrm{MLP}}\|_{F}^{2}$.
By contrast, the down step involves a design matrix of size $T \times D_{\mathrm{up}}$, so the associated normal matrix is only $D_{\mathrm{up}} \times D_{\mathrm{up}}$ and can be handled explicitly. This allows the corresponding least-squares problem to be solved in closed form.

\section{Additional Theoretical Results}
\label{sec:appendix_theory}

\subsection{QEP for Activation Quantization}
\label{subsec:QEP-activation}

In this subsection, we develop a QEP-style extension of activation quantization that was not addressed in the original QEP \citep{arai2025quantization}, and we demonstrate that it emerges naturally as a single LPCD update on the activation block.

When both weights and activations are quantized, several design patterns are commonly used in practice.
One approach quantizes weights and activations independently, typically through direct approximation. 
A second approach first quantizes the activations and then optimizes the weights based on an activation-aware objective defined in terms of the quantized activations.
A third approach jointly optimizes the weight and activation quantizers to directly minimize output-level discrepancy, often employing one of the previous strategies for initialization.
The LPCD offers a natural method to enhance the second and third approaches by deriving a QEP-style update for the activations, analogous to the original QEP update for the weights.
Recall the two-block objective used in the weight-side analysis in Section~\ref{subsec:qep-lpcd}:
\begin{equation}
  \label{eq:qep-lpcd-obj-activation}
  L(\widehat{W},\widehat{X})
  = \|\widehat{X}\widehat{W} - XW \|_{F}^{2},
\end{equation}
In the activation-side variant, the weight block $\widehat{W}$ is treated as fixed, while the LPCD step is applied to the activation block $\widehat{X}$.

\paragraph{Relaxation Step.}
The relaxation step can be expressed as
\begin{equation}
  \label{eq:qep-activation-step1-problem}
  \overline{X}^{(1)}
  = \argmin_{U}
  \|U\widehat{W} - XW \|_{F}^{2}.
\end{equation}
A standard matrix calculus computation demonstrates that the gradient is
\begin{equation}
  \nabla_{U} \|U\widehat{W}-XW\|_{F}^{2} =
  2 (U\widehat{W} - XW)\widehat{W}^\top.
\end{equation}
Any minimizer $\widehat{X}^\ast$ of Eq.~\eqref{eq:qep-activation-step1-problem} satisfies the first-order optimality condition:
\begin{equation}
  \label{eq:qep-activation-first-order}  (\widehat{X}^{\ast}\widehat{W} - XW)\widehat{W}^{\top} = 0.
\end{equation}
Assume the matrix $\widehat{W}\widehat{W}^{\top}\in \mab{R}^{N \times N}$ is invertible.
In this case, the unique solution of Eq.~\eqref{eq:qep-activation-step1-problem} is
$\widehat{X}^{\ast} = XW\widehat{W}^\top (\widehat{W}\widehat{W}^{\top})^{-1}$. 
If $\widehat{W}\widehat{W}^{\top}$ is not invertible, the same derivation shows that any minimizer can be expressed as
\begin{equation}
  \overline{X}^{(1)}
  =
  XW\widehat{W}^{\top}
  (\widehat{W}\widehat{W}^\top)^{\dagger} +
  Z(I_{N} - \widehat{W}\widehat{W}^\top(\widehat{W}\widehat{W}^\top)^{\dagger}),
\end{equation}
for some $Z\in\mab{R}^{T\times N}$, where $(\cdot)^{\dagger}$ denotes the Moore–Penrose pseudoinverse. 
The choice $Z=0$ yields the minimum-norm least-squares solution $\overline{\widehat{X}}^{(1)} = XW\widehat{W}^{\top}(\widehat{W}\widehat{W}^{\top})^{\dagger}$, which coincides with the solution obtained when $\widehat{W}\widehat{W}^\top$ is invertible.
In either case, the continuous LPCD relaxation produces a \emph{corrected} activation matrix $\overline{\widehat{X}}^{(1)}$ that best aligns the quantized layer output $U\widehat{W}$ with the full-precision output $XW$ in the least-squares sense.

\paragraph{Projection Step.}
The second LPCD step enforces the activation quantization constraint by projecting $\overline{\widehat{X}}^{(1)}$ onto the set $\mab{Q}^{T\times N}$ using the direct projector $\Pi_{\mab{Q}}^{(d)}$.
Applying this projection to $\widehat{X}^\ast$ results in the activation-side QEP update
\begin{equation}
  \label{eq:qep-activation-update}
  \widehat{X}^{\mathrm{QEP}} = \Pi_{\mab{Q}}^{(d)}(\overline{X}^{(1)}) = \Pi_{\mab{Q}}^{(d)}
  \ab(XW\widehat{W}^{\top}(\widehat{W}\widehat{W}^{\top})^{-1}).
\end{equation}
Eq.~\eqref{eq:qep-activation-step1-problem}--\eqref{eq:qep-activation-update} demonstrates that extending QEP from weights to activations involves applying a single LPCD update to the activation block (the block with index $r=2$ in the two-block objective Eq.~\eqref{eq:qep-lpcd-obj-activation}), while the weight block $\widehat{W}$ remains fixed.
This construction yields a mathematically consistent activation analog of QEP and integrates both the weight-side and activation-side QEP within the unified LPCD.

\subsection{QEP for KV-Cache Quantization}\label{subsec:qep-kv-cache}

This section extends QEP and LPCD to KV-cache quantization.
We demonstrate that both the key and value caches allow for QEP-style corrections, which can be interpreted as single LPCD updates on their respective activation blocks, followed by standard activation quantizers.

\paragraph{Setting.}
We consider a single Transformer layer and a single attention head for clarity; the multi-head case follows by applying the same derivations to each head independently. Let
\begin{equation}
    Q \in \mab{R}^{T_{q} \times d_{k}}, ~~~
    K \in \mab{R}^{T_{k} \times d_{k}}, ~~~
    V \in \mab{R}^{T_{k} \times d_{v}}
\end{equation}
denote the full-precision query, key, and value matrices, respectively.
To simplify, we ignore the scaling factor and causal masking. 
These can be absorbed into $Q$ and $K$, as shown in Appendix~\ref{subsub:kq-module}. the full-precision attention logits and outputs are
\begin{equation}
    S = QK^{\top} \in \mab{R}^{T_q \times T_k},~~
    A = \operatorname{Softmax}(S),~~Y = AV \in \mab{R}^{T_q \times d_v},
\end{equation}
where the softmax is applied row-wise.
At inference time, we store a quantized KV cache $(\widehat{K}, \widehat{V})$ and use quantized queries $\widehat{Q}$.
Our goal is to adjust $\widehat{K}$ and $\widehat{V}$ within LPCD while preserving a standard activation quantization interface.

\subsubsection{Kye-Cache Update}
We derive a QEP-style update for the key cache.
Following the QEP design principle, we align the \emph{pre-softmax} attention scores computed from the quantized path with their full-precision counterparts.
We therefore minimize the Frobenius norm
\begin{equation}
    L_{K}(\widehat{K})
    = \|\widehat{Q}\widehat{K}^\top - QK^{\top} \|_{F}^{2},
    \label{eq:qep-k-key-loss}
\end{equation}
where $\widehat{Q}$ denotes the quantized query matrix produced by the preceding submodule.
This corresponds to the block-wise objective of LPCD with the key block treated as an activation block.

\paragraph{Relaxation Step.}
Let $U = \widehat{K}^\top \in \mab{R}^{d_k \times T_k}$ represent the relaxed variable.
The objective Eq.~\eqref{eq:qep-k-key-loss} is defined as
\begin{equation}
    L_{K}(U)
    = \|\widehat{Q}U - S\|_{F}^{2},~~
    S = QK^\top.
\end{equation}
The optimality condition for this least-squares problem is as follows:
\begin{equation}
    \widehat{Q}^{\top}(\widehat{Q}U - S) = 0.
\end{equation}
Defining $\widehat{H}_{Q} = \widehat{Q}^\top\widehat{Q}$ and $    C_{Q} = \widehat{Q}^\top(Q-\widehat{Q})$. We obtain
\begin{equation}
    \widehat{H}_{Q} U
    = \widehat{Q}^\top S
    = \widehat{Q}^\top QK^\top
    = (\widehat{H}_{Q} + C_{Q})K^{\top}.
\end{equation}
Assuming that $\widehat{H}_{Q}$ is invertible or replacing it with a regularized inverse, the unique minimizer is
\begin{equation}
    U^\ast =
    \widehat{H}_{Q}^{-1}(\widehat{H}_{Q} + C_{Q})K^\top
    = (I + \widehat{H}_{Q}^{-1}C_{Q})K^{\top}.
\end{equation}
Transposing back to the original parameterization yields the relaxed key cache
\begin{equation}
    \overline{K}^{(1)}
    =
    (U^\ast)^\top
    =
    K\ab(I + \widehat{H}_{Q}^{-1}C_{Q})^\top.
    \label{eq:qep-k-key-relaxed}
\end{equation}
As in QEP for weights, we introduce a tunable coefficient $\alpha_{K}\in[0,1]$ to interpolate between the original and fully corrected keys:
\begin{equation}
    \overline{K}^{(1)}(\alpha_{K})
    =
    K\ab(I + \alpha_{K}\widehat{H}_{Q}^{-1}C_{Q})^\top.
    \label{eq:qep-k-key-relaxed-alpha}
\end{equation}
When $\alpha_{K} = 0$, we obtain $\overline{K}^{(1)} = K$; when $\alpha_{K} = 1$, we recover the full least-squares solution.

\paragraph{Projection Step.}
Within our LPCD formulation, the key cache is treated as an activation block, i.e., $r \in \mathcal{R}_{a}$.
Thus, the projection step applies the direct activation quantizer $\Pi_{\mab{Q}}^{(d)}$ to the relaxed keys:
\begin{equation}
    \widehat{K}^{(1)}
    =
    \Pi_{\mab{Q}}^{(d)}\ab(\overline{K}^{(1)}(\alpha_{K})),
    \label{eq:qep-k-key-proj}
\end{equation}
where $\Pi_{\mab{Q}}^{(d)}$ is instantiated using either per-channel or per-token schemes (e.g., KIVI-style asymmetric quantization).
Eq.~\eqref{eq:qep-k-key-relaxed-alpha}--\eqref{eq:qep-k-key-proj} thus defines a single LPCD update for the key-cache block with a QEP-style correction.

\subsubsection{Value-Cache Update}

Next, we derive a QEP-style update for the value cache.
Here, we align the \emph{post-softmax} attention outputs by using the quantized attention weights as the design matrix.
Let $\widehat{S} = \widehat{Q}\widehat{K}^\top$ and $\widehat{A} = \operatorname{Softmax}(\widehat{S})$
be the (fixed) logits and attention weights computed from the quantized queries and keys.
We define the loss
\begin{equation}
    L_{V}(\widehat{V})
    =
    \ab\|
        \widehat{A}\widehat{V} - AV
    \|_{F}^{2}
    =
    \ab\|
        \widehat{A}\widehat{V} - Y
    \|_{F}^{2},
    \label{eq:qep-k-value-loss}
\end{equation}
which again matches the QEP pattern.

\paragraph{Relaxation Step.}
Let $U = \widehat{V} \in \mab{R}^{T_k \times d_v}$ denote the relaxed variable.
The objective expressed in Eq.~\eqref{eq:qep-k-value-loss} can be stated as
\begin{equation}
    L_{V}(U)
    =
    \|\widehat{A}U - Y\|_{F}^{2},~~
    Y = AV.
\end{equation}
The normal equations are:
\begin{equation}
    \widehat{A}^\top\ab(\widehat{A}U - Y) = 0.
\end{equation}
By defining $\widehat{H}_{A} = \widehat{A}^\top\widehat{A}$ and $C_{A} = \widehat{A}^\top(A - \widehat{A})$,
we obtain
\begin{equation}
    \widehat{H}_{A} U
    = \widehat{A}^\top Y
    = \widehat{A}^\top AV
    = (\widehat{H}_{A} + C_{A})V.
\end{equation}
Assuming $\widehat{H}_{A}$ is invertible, the least-squares minimizer is
\begin{equation}
    U^\ast
    =
    \widehat{H}_{A}^{-1}(\widehat{H}_{A} + C_{A})V
    =
    \ab(I + \widehat{H}_{A}^{-1}C_{A})V.
\end{equation}
Thus, the relaxed value cache is
\begin{equation}
    \overline{V}^{(1)}
    =
    \ab(I + \widehat{H}_{A}^{-1}C_{A})V.
    \label{eq:qep-k-value-relaxed}
\end{equation}
Introducing a propagation strength parameter $\alpha_{V}\in[0,1]$ yields
\begin{equation}
    \overline{V}^{(1)}(\alpha_{V})
    =
    \ab(I + \alpha_{V}\widehat{H}_{A}^{-1}C_{A})V,
    \label{eq:qep-k-value-relaxed-alpha}
\end{equation}
which interpolates between the original values and the complete solution.

\paragraph{Projection Step.}
Regarding the key cache, the value cache is considered an activation block in our LPCD.
We therefore apply the direct activation projection:
\begin{equation}
    \widehat{V}^{(1)}
    =
    \Pi_{\mab{Q}}^{(d)}\ab(\overline{V}^{(1)}(\alpha_{V})).
    \label{eq:qep-k-value-proj}
\end{equation}
In practice, per-token schemes are often preferable for values, whereas keys benefit from per-channel schemes; our formulation is agnostic to this choice and simply reuses the underlying activation quantizer.

Eqs.~\eqref{eq:qep-k-key-loss}--\eqref{eq:qep-k-value-proj} demonstrate that the QEP for KV caches is derived by executing a single LPCD update on the key and value-cache blocks with suitably chosen block-wise objectives.
The key update aligns pre-softmax attention logits, while the value update aligns post-softmax attention outputs.
Both updates are expressed in closed form during the relaxation step and then projected using standard activation quantizers, thereby preserving compatibility with existing KV-cache quantization schemes, such as per-channel and per-token uniform quantization.
In Sec.~\ref{subsec:submodule-wise-ptq}, we use these KV-cache updates as building blocks for submodule LPCD applied to the grouped QK and VO modules.

\subsection{QEP for Orthogonal Rotation Matrices}
\label{subsec:lpcd-rotation}

Rotation-based incoherence processing~\citep{frantar2022gptq, tseng2024quip, liu2024spinquant} requires a linear map represented by an orthogonal matrix $R \in \mab{R}^{N \times N}$ to redistribute outliers across channels. 
For a weight matrix
$W \in \mab{R}^{N \times M}$ and activations
$X \in \mab{R}^{T \times N}$,
the transformation
\begin{equation}
    Y = XW = (XR)(R^\top W)
    \label{eq:rotation-factorization}
\end{equation}
leaves the full-precision output unchanged while operating in the
channel dimension.
In rotation-aware PTQ, one typically quantizes the rotated quantities $XR$ and $R^{\top} W$;
their quantized counterparts are referred to as $\widehat{X} \in \mab{R}^{T \times N}$ and
$\widehat{W} \in \mab{R}^{N \times M}$, respectively.
In this section, we treat the rotation matrix $R$ as a block in LPCD and derive a single update for $R$ while keeping $\widehat{X}$ and $\widehat{W}$ fixed. 
Compared to optimization methods such as CalySGD introduced in SpinQuant, the QEP-style updates described below are more memory-efficient, as only the linear layers need to be stored in memory. Additionally, it provides more stable optimization than CalySGD.

\paragraph{Objective.}
Motivated by the factorization in Eq.~\eqref{eq:rotation-factorization},
we select $R$ so that the rotated full-precision tensors
$XR$ and $R^{\top}W$ are well aligned with their corresponding fixed quantized
counterparts $\widehat{X}$ and $\widehat{W}$.
We therefore consider the quadratic objective
\begin{equation}
    L_{R}(R)
    =
    \lambda_{a}
    \|XR - \widehat{X}\|_{F}^{2}
    +
    \lambda_{w}
    \|R^{\top} W - \widehat{W}\|_{F}^{2},
    \label{eq:rotation-loss}
\end{equation}
where $\lambda_{a}, \lambda_{w} \ge 0$ balances the contributions of activation
and weight. 
This defines the LPCD block-loss for the rotation block
$M_{r} = R$ while keeping all other blocks fixed.

\paragraph{Relaxation Step.}
We relax the orthogonality constraint and minimize Eq.~
\eqref{eq:rotation-loss} over all real matrices
$R \in \mab{R}^{N \times N}$.
The first term in Eq.~\eqref{eq:rotation-loss} can be expressed as
\begin{equation}
    f_{a}(R)
    =
    \|XR - \widehat{X}\|_{F}^{2}
    =
    \operatorname{tr}\ab[(XR-\widehat{X})^\top (XR-\widehat{X})].
\end{equation}
Its gradient is
\begin{equation}
    \nabla_{R} f_{a}(R)
    =
    2X^\top(XR-\widehat{X}).
\end{equation}
For the second term,
\begin{equation}
    f_{w}(R)
    =
    \|R^{\top} W - \widehat{W}\|_{F}^{2}
    =
    \|W^\top R - \widehat{W}^{\top}\|_{F}^{2},
\end{equation}
we obtain
\begin{equation}
    \nabla_{R} f_{w}(R)
    =
    2W(W^{\top} R - \widehat{W}^\top)
    =
    2(WW^\top R - W\widehat{W}^\top).
\end{equation}
Combining both contributions, the total gradient of
$L_{R}(R)$ is expressed as
\begin{equation}
    \nabla_{R} L_{R}(R)
    =
    2\lambda_{a} X^\top(XR-\widehat{X})
    +
    2\lambda_{w} (WW^\top R - W\widehat{W}^\top).
\end{equation}
Setting the gradient to zero yields the normal equations
\begin{equation}
    (\lambda_{a} X^\top X + \lambda_{w} WW^\top ) R
    =
    \lambda_{a} X^\top \widehat{X}
    +
    \lambda_{w} W\widehat{W}^\top.
    \label{eq:rotation-normal-equation}
\end{equation}
We define $H_{R}=\lambda_{a} X^{\top} X + \lambda_{w} WW^{\top}$ and $B_{R}=\lambda_{a} X^\top \widehat{X}+ \lambda_{w} W\widehat{W}^\top.$
Assuming $H_{R}$ is invertible or that a regularized inverse is used,
the relaxed minimizer can be expressed in closed form as follows:
\begin{equation}
    \overline{R}^{(1)}
    =
    H_{R}^{-1} B_{R}.
    \label{eq:rotation-relaxed-solution}
\end{equation}
This step corresponds to the LPCD relaxation for the
rotation block: we compute the continuous matrix
$\overline{R}^{(1)}$ that satisfies both the activation and weight constraints in the least-squares sense.

\paragraph{Projection Step.}
The rotation matrix is required to be orthogonal,
\begin{equation}
    \mathcal{O}(N)
    =
    \{R \in \mab{R}^{N \times N} : R^{\top} R = I_{N}\}.
\end{equation}
To restore this constraint, we project the relaxed solution
$\overline{R}^{(1)}$ onto $\mathcal{O}(N)$ using the Frobenius norm:
\begin{equation}
    R^{(1)}
    =
    \Pi_{\mathcal{O}(N)}\ab(\overline{R}^{(1)})
    \coloneqq
    \argmin_{R \in \mathcal{O}(N)}
    \ab\| R - \overline{R}^{(1)} \|_{F}^{2}.
    \label{eq:rotation-projection-problem}
\end{equation}
This is the classical orthogonal Procrustes problem.
Let the singular value decomposition of
$\overline{R}^{(1)}$ be
\begin{equation}
    \overline{R}^{(1)}
    =
    U\Sigma V^\top,
\end{equation}
with $U,V \in \mab{R}^{N \times N}$ orthogonal and
$\Sigma$ diagonal matrices that have nonnegative entries.
Then the unique minimizer of Eq.~
\eqref{eq:rotation-projection-problem} is
\begin{equation}
    R^{(1)}
    =
    U V^\top.
    \label{eq:rotation-projection-solution}
\end{equation}
If one wishes to restrict to proper rotations, $\det(R^{(1)}) = 1$), the sign of the last column of $U$
or $V$ can be flipped accordingly.
Eqs.~~\eqref{eq:rotation-relaxed-solution} and
\eqref{eq:rotation-projection-solution} collectively define a single LPCD
update for the rotation block with fixed quantized
$\widehat{X}$ and $\widehat{W}$.

\subsection{QEP for LoRA-Based Error Compensation}
\label{subsec:QEP-LoRA}

In this section, we extend the LPCD to a setting in which the quantization error is compensated by a low-rank adapter that follows the LoRA framework~\citep{hu2022lora, dettmers2023qlora}.
We treat the LoRA correction as a separate block variable and describe its update solely in terms of the relaxation and projection steps of LPCD.

\paragraph{LoRA Parameterization.}
Suppose we have a \emph{base} quantized weight $\widehat{W}_{0} \in \mab{Q}^{N\times M}$ obtained using a layer-wise PTQ method such as GPTQ or AWQ.
LoRA parameterizes an additive correction as follows:
\begin{equation}
  \widehat{W} = \widehat{W}_{0} + \Delta W,~~~\Delta W = B A,
\end{equation}
where $B\in\mab{R}^{N\times r}$, $A \in \mab{R}^{r\times M}$, and $r \ll \min\{N, M\}$.
Thus, any LoRA-style model corresponds to a weight matrix within the affine set
\begin{equation}
  \widehat{W} \in \widehat{W}_{0} + \mac{M}_{r},~~~\mac{M}_{r} \coloneqq \ab\{E\in\mab{R}^{N\times M} : \mathrm{rank}(E)\le r\}.
\end{equation}
We consider the following two-block objective:
\begin{equation}
  \label{eq:qep-lora-lpcd-objective}
  L(\widehat{W}_{0},E)
  \coloneqq
  \ab\|\widehat{X}(\widehat{W}_{0}+E) - XW\|_{F}^{2},
\end{equation}
with blocks $M_{1}=\widehat{W}_{0}\in\mab{Q}^{N\times M}$, the base quantized weight, and $M_{2}=E\in\mab{R}^{N\times M}$, the LoRA correction.
We fix $M_{1}=\widehat{W}_{0}$ and apply one LPCD update to $M_{2}$.

\paragraph{Relaxation Step.}
Define the output residual of the base quantized weight
\begin{equation}
  R \coloneqq XW - \widehat{X}\widehat{W}_{0} \in \mab{R}^{T\times M},
\end{equation}
so that
\begin{equation}
  L(\widehat{W}_{0},E) = \|\widehat{X}E - R\|_{F}^{2}.
\end{equation}
The relaxation step minimizes $L$ over $E$ without imposing any rank constraints:
\begin{equation}
  \overline{E}^{(1)}
  = \argmin_{U\in\mab{R}^{N\times M}} \|\widehat{X}U - R\|_{F}^{2}.
\end{equation}
Let $\widehat{H}\coloneqq \widehat{X}^{\top}\widehat{X}\in\mab{R}^{N\times N}$.
The first-order optimality condition is expressed as follows:
\begin{equation}
  \widehat{H}\overline{E}^{(1)} = \widehat{X}^{\top}R.
\end{equation}
If $\widehat{H}$ is invertible, the unique minimizer is given by
\begin{equation}
  \label{eq:qep-lora-relax-closed-form}
  \overline{E}^{(1)}
  =
  \widehat{H}^{-1}\widehat{X}^{\top}R
  = \widehat{H}^{-1}\widehat{X}^{\top}(XW-\widehat{X}\widehat{W}_{0}).
\end{equation}
In general, we utilize the Moore–Penrose pseudoinverse and define
\begin{equation}
  \label{eq:qep-lora-relax-pinv}
  \overline{E}^{(1)}
  =
  \widehat{H}^{\dagger}\widehat{X}^{\top}R,
\end{equation}
which represents the minimum-norm least-squares solution.

\paragraph{Projection Step.}
The projection step enforces the LoRA rank constraint by projecting $\overline{E}^{(1)}$ onto the set $\mac{M}_{r}$ in a manner that aligns with the activation-aware metric.
We use the $\widehat{H}$-weighted norm
\begin{equation}
  \label{eq:qep-lora-weighted-norm}
  \|E\|_{\widehat{H}}^{2}
  \coloneqq
  \mathrm{tr}(E^{\top}\widehat{H}E)
  = \|\widehat{H}^{1/2}E\|_{F}^{2}.
\end{equation}
We define the following projection:
\begin{equation}
  E^{(1)}
  =
  \Pi^{(\mathrm{LoRA})}_{r}(\overline{E}^{(1)})
  \coloneqq
  \argmin_{E:\mathrm{rank}(E)\le r}
  \|E-\overline{E}^{(1)}\|_{\widehat{H}}^{2}.
\end{equation}
Let $F^{(1)} \coloneqq \widehat{H}^{1/2}\overline{E}^{(1)} \in \mab{R}^{N\times M}$ and compute its singular value decomposition $F^{(1)} = U\Sigma V^{\top}$.
Denote by $U_{r}\in\mab{R}^{N\times r}$, $\Sigma_{r}\in\mab{R}^{r\times r}$, and $V_{r}\in\mab{R}^{M\times r}$ the matrices obtained by truncating to the top $r$ singular values.
The weighted best rank-$r$ approximation of $\overline{E}^{(1)}$ is given by
\begin{equation}
  \label{eq:qep-lora-proj-closed-form}
  E^{(1)}
  =
  \widehat{H}^{-1/2} U_{r}\Sigma_{r}V_{r}^{\top}.
\end{equation}
The updated weight after one LPCD iteration on the LoRA block is
\begin{equation}
  \widehat{W}^{(1)}
  =
  \widehat{W}_{0} + E^{(1)}.
\end{equation}
If an explicit LoRA factorization is required, we define $B \coloneqq \widehat{H}^{-1/2}U_{r}\Sigma_{r}^{1/2}$ and $A \coloneqq \Sigma_{r}^{1/2}V_{r}^{\top}$,
such that $BA = E^{(1)}$ and $\widehat{W}^{(1)} = \widehat{W}_{0} + BA$.
In this manner, error compensation using LoRA is achieved through a single LPCD update on the low-rank correction block, without altering the base layer-wise PTQ operator.

\end{document}